\documentclass{article}

\usepackage[utf8]{inputenc} 
\usepackage{microtype}
\usepackage{graphicx}
\usepackage{fontawesome}
\usepackage{subfigure}
\usepackage{booktabs} 

\usepackage{hyperref}

\usepackage[preprint]{icml2026}

\usepackage{amsmath}
\usepackage{amssymb}
\usepackage{mathtools}
\usepackage{amsthm}
\usepackage{amsfonts}
\usepackage{nicefrac}       

\usepackage{url}            
\usepackage{color,xcolor}   
\usepackage{caption}
\usepackage{subcaption}
\usepackage{setspace}
\usepackage{multirow}
\usepackage{threeparttable}
\usepackage{makecell}
\usepackage{pifont}
\usepackage{xspace}
\usepackage[misc]{ifsym}
\usepackage{colortbl}
\usepackage{wrapfig}

\newcolumntype{K}[1]{>{\centering\arraybackslash}p{#1}}

\hypersetup{
    pdftitle={},
    pdfsubject={},
    pdfauthor={},
    pdfkeywords={},
}

\usepackage[capitalize,noabbrev]{cleveref}

\theoremstyle{plain}

\theoremstyle{definition}

\theoremstyle{remark}

\usepackage[textsize=tiny]{todonotes}

\newcommand{\modelname}{\textsc{MoHETS}}
\newcommand{\model}{\modelname\xspace}
\newcommand{\tmodel}{\modelname\textsubscript{tiny}\xspace}
\newcommand{\smodel}{\modelname\textsubscript{small}\xspace}
\newcommand{\basemodel}{\modelname\textsubscript{base}\xspace}
\newcommand{\largemodel}{\modelname\textsubscript{large}\xspace}
\newcommand{\dmodel}{$d_{\text{model}}$\xspace}
\newcommand{\dff}{$d_{\text{ff}}$\xspace}

\newcommand{\bestres}[1]{{\textbf{\textcolor{red}{#1}}}}
\newcommand{\secondres}[1]{{\underline{\textcolor{blue}{#1}}}}

\definecolor{fbApp}{HTML}{ffe4e3}
\definecolor{tabhighlight}{HTML}{e5e5e5}

\def\Figref#1{Figure~\ref{#1}}

\def\Tabref#1{Table~\ref{#1}}

\def\eqref#1{equation~\ref{#1}}

\icmltitlerunning{MoHETS: Long-term Time Series Forecasting with Mixture-of-Heterogeneous-Experts}

\begin{document}

\twocolumn[
\icmltitle{\modelname: Long-term Time Series Forecasting with\\ Mixture-of-Heterogeneous-Experts}



\icmlsetsymbol{equal}{*}

\begin{icmlauthorlist}

\icmlauthor{Evandro S. Ortigossa}{wis}
\icmlauthor{Guy Lutsker}{wis,mcb}
\icmlauthor{Eran Segal}{wis,mbzu}


\end{icmlauthorlist}

\begin{center}
\faGithub~\footnotesize{Code is available at \url{https://github.com/evortigosa/mohe_forecast}}
\end{center}

\icmlaffiliation{wis}{Department of Computer Science and Applied Mathematics, Weizmann Institute of Science, Rehovot, Israel}
\icmlaffiliation{mcb}{Department of Molecular Cell Biology, Weizmann Institute of Science, Rehovot, Israel}
\icmlaffiliation{mbzu}{Mohamed bin Zayed University of Artificial Intelligence, Abu Dhabi, UAE}

\icmlcorrespondingauthor{Eran Segal}{eran.segal@weizmann.ac.il}

\icmlkeywords{Machine Learning, Deep Learning, ICML, Mixture-of-Experts, Forecasting, Transformers}

\vskip 0.3in
]



\printAffiliationsAndNotice{}  

\begin{abstract}
    Real-world multivariate time series can exhibit intricate multi-scale structures, including global trends, local periodicities, and non-stationary regimes, which makes long-horizon forecasting challenging. Although sparse Mixture-of-Experts (MoE) approaches improve scalability and specialization, they typically rely on homogeneous MLP experts that poorly capture the diverse temporal dynamics of time series data. We address these limitations with MoHETS, an encoder-only Transformer that integrates sparse Mixture-of-Heterogeneous-Experts (MoHE) layers. MoHE routes temporal patches to a small subset of expert networks, combining a shared depthwise-convolution expert for sequence-level continuity with routed Fourier-based experts for patch-level periodic structures. MoHETS further improves robustness to non-stationary dynamics by incorporating exogenous information via cross-attention over covariate patch embeddings. Finally, we replace parameter-heavy linear projection heads with a lightweight convolutional patch decoder, improving parameter efficiency, reducing training instability, and allowing a single model to generalize across arbitrary forecast horizons. We validate across seven multivariate benchmarks and multiple horizons, with MoHETS consistently achieving state-of-the-art performance, reducing the average MSE by $12\%$ compared to strong recent baselines, demonstrating effective heterogeneous specialization for long-term forecasting.
\end{abstract}

\section{Introduction}
\label{sec:intro}

    Time series forecasting is a critical task for decision-making in a wide variety of domains such as energy management~\cite{lago2021forecasting}, financial planning~\cite{nie2024survey}, healthcare~\cite{lutsker2025glucose}, and climate analysis~\cite{zhang2023skilful}. However, accurately predicting future values from historical observations is challenging, as real‐world time series data often present complex temporal dependencies, seasonality, trends, non-stationarity, and exogenous influences that lead to varied distributions even within short context windows~\cite{wang2024timexer,liu2024moirai}. 
    Traditional statistical methods, including ARIMA, exponential smoothing, and vector autoregression~\cite{ortigossa2025time}, often struggle to capture nonlinear patterns, multiple seasonalities, or high-dimensional multivariate interactions. 
    Furthermore, these challenges intensify in long-term time series forecasting, where predicting intricate cross-variate dependencies over extended contexts demands models that scale efficiently while maintaining predictive accuracy~\cite{liu2023itransformer}.

    Deep learning approaches have demonstrated remarkable performance in time series forecasting by enabling robust modeling of nonlinear and multivariate patterns. Initially designed for natural language processing (NLP), the Transformer architecture~\cite{vaswani2017attention} has been successfully extended to computer vision (CV)~\cite{dosovitskiy2020image}, audio~\cite{gong2021ast}, and time series~\cite{zhou2021informer,wen2023transformers}. Transformer models introduce attention mechanisms that adaptively weight historical information to capture long-range dependencies~\cite{zhou2021informer}.
    However, a fundamental misalignment persists in the context of time series: standard Transformers, designed for the discrete semantics of NLP, apply homogeneous processing, typically using dense MLPs, to all tokens. Time series data, conversely, are composed of distinct structural components, persistent periodicities, and transient trends that require fundamentally different inductive biases. Applying a uniform architecture to disentangle these heterogeneous patterns often results in inefficient parameter usage and suboptimal fittings~\cite{dong2024fan}.

    Recent improvements have pushed Transformers to mitigate these issues. Patching techniques reduce complexity~\cite{nie2022time}, integration of exogenous variables allows learning contextual correlations~\cite{liu2024timer,wang2024timexer}, and sparse designs enable more efficient scaling~\cite{shi2024time,liu2024moirai}. Although recent sparse approaches have introduced Mixture-of-Experts (MoE) to reduce computational overhead, they essentially inherit the homogeneous expert design of large language models (LLMs), where every expert is an identical MLP~\cite{shi2024time,liu2024moirai}. This design ignores the multi-scale nature of time series, where capturing high-frequency local variations requires different operators than modeling long-term global dependencies. 
    To address these challenges, we designed \model, a novel encoder-only Transformer model that integrates sparse Mixture-of-Heterogeneous-Experts ($\operatorname{MoHE}$) with patch-based embedding and covariate integration, improving its robustness to diverse temporal dependencies and patterns. 
    We demonstrate that \model consistently outperforms well-known state-of-the-art models in forecasting benchmark experiments.
    In summary, the main contributions of this research are as follows:
    \begin{itemize}
        \item We introduce \model 
        , an encoder-only Transformer with a Mixture-of-Heterogeneous-Experts ($\operatorname{MoHE}$) strategy that applies architecturally distinct experts to effectively model time patterns at different levels, ensuring the model architecture aligns with the intrinsic decomposition of time-series data.

        \item We incorporate a multimodal cross-attention module that integrates external information from exogenous covariates. With this design, \model enhances time series representations by capturing interactions between endogenous features and exogenous information.

        \item We propose the $\operatorname{MoHE}$ layer, which combines depthwise convolutions and Fourier-based experts to capture global trends and local periodicities at the patch level, respectively, enhancing specialization while maintaining the scaling benefits of standard $\operatorname{MoEs}$.
    \end{itemize}
    \vspace{1pt}

\section{Related Work}
\label{sec:related}

\subsection{Deep Learning for Time Series Forecasting}
\label{subsec:ts_dl}

    Deep learning models have significantly advanced time series forecasting, transitioning from MLP-based networks~\cite{wangtimemixer}, recurrent neural networks (RNNs)~\cite{salinas2020deepar}, and convolutional neural networks (CNNs)~\cite{sen2019think} to Transformer-based architectures~\cite{wen2023transformers}. 
    The attention mechanism of Transformers is able to adaptively weight historical information, making it a natural choice for handling long-term dependencies. Early Transformer models, such as Informer~\cite{zhou2021informer}, introduced the ProbSparse self-attention to address the quadratic complexity of standard Transformers, while Autoformer~\cite{wu2021autoformer} leveraged auto-correlation to discover period-based dependencies at the subseries level. 
    Recently, PatchTST~\cite{nie2022time} further improved efficiency by applying channel-independent processing and segmenting time series into subseries-level patches, reducing computational overhead while preserving local semantics and allowing the model to attend longer context windows. 
    iTransformer~\cite{liu2023itransformer} inverts the input dimensions to apply attention across variates rather than time, prioritizing multivariate correlations over temporal dependencies. Foundation models, such as TimeGPT~\cite{garza2023timegpt} and TimesFM~\cite{das2023decoder}, explore pre-training paradigms to improve adaptability. However, these architectures are predominantly dense and homogeneous. By processing diverse temporal dynamics, such as high-frequency noise and low-frequency trends, through identical operators (e.g., standard MLPs), they suffer from parameter redundancy and struggle to decouple entangled temporal patterns.

\subsection{Forecasting with Covariates}
\label{subsec:covs}

    Real-world time series are often partially observed, and endogenous variables (the primary series to forecast) are frequently influenced by exogenous covariates, which capture external contexts that can affect temporal dynamics and consequently predictions~\cite{wang2024timexer}. Covariates encompass external contexts such as calendar events, weather metrics, or economic indicators that drive the non-stationary dynamics of the target series. 
    Transformer-based models have increasingly introduced covariates to improve contextual understanding. The Temporal Fusion Transformer (TFT)~\cite{lim2021temporal} employed variable selection networks and entity embeddings to dynamically weigh covariates, while Timer-XL~\cite{liu2024timerxl} supports covariate-informed forecasting in a decoder-only patched architecture.
    Incorporating such data requires careful handling to address issues such as missing values or temporal misalignment. In this context, TimeXer~\cite{wang2024timexer} refined covariate incorporation by implementing a patch-wise self-attention module for endogenous series and a variate-wise cross-attention module for exogenous inputs, thus mitigating issues related to partial observability and temporal misalignment. 
    Exogenous covariates can provide valuable external information to enhance robustness to non-stationarity and improve forecasting accuracy. However, current methods often append covariates as auxiliary tokens or simplified concatenations~\cite{lim2021temporal,das2023long}, failing to explicitly model the cross-modal interaction between static external contexts and dynamic time patches.

\subsection{Sparse Mixture-of-Experts (MoE)}
\label{subsec:moe}

    Deep learning models are dense, imposing high memory and computational costs during training and inference~\cite{shi2024time}. Sparse architectures, such as Mixture-of-Experts ($\operatorname{MoE}$), dynamically route inputs to specialized sub-networks for conditional activation, enabling scaling up the model's capacities while reducing computational overhead~\cite{shazeer2017outrageously,fedus2022switch}. Model sparsification has received considerable attention in the context of NLP and CV for efficient handling of diverse patterns~\cite{fedus2022switch,mixtralmoe,dai2024deepseekmoe,riquelme2021scaling}, but has received relatively less attention in time-series research, with few relevant works implementing sparse approaches~\cite{shi2024time}. 
    MoLE~\cite{ni2024mixture} explores linear ensembling, Time-MoE~\cite{shi2024time} and Moirai-MoE~\cite{liu2024moirai} adapt sparse $\operatorname{MoEs}$ to decoder-only Transformers, while Seg-MoE \cite{ortigossa2026seg} proposes a $\operatorname{MoE}$ routing design that processes contiguous time-step segments rather than independent tokens.
    However, these models strictly adhere to the NLP-standard and rely on MLP-based expert designs. 
    Training stability and expert specialization are challenges when applying MoE, as routing mechanisms can lead to load imbalance or overfitting~\cite{fedus2022switch}. 
    Crucially, the direct adaptation of NLP-centric $\operatorname{MoEs}$ overlooks the signal processing nature of time series. Standard MLP experts lack the inductive bias to efficiently separate global trends from local periodicities~\cite{dong2024fan}, a task for which specialized operators such as Convolutions and Fourier Transforms have been mathematically shown to outperform MLPs~\cite{wangtimemixer}.
    
    We bridge this gap with \model, replacing homogeneous MLPs with a Mixture-of-Heterogeneous-Experts that assigns architecturally distinct operators to the specific temporal components they are best suited to model.

\section{Methodology}
\label{sec:methodology}

    \vspace{0.1cm}\textbf{Problem Statement.} Let $\mathbf{X} \in \mathbb{R}^{D \times T}$ denote a set of multivariate time series with $D$ variates (or channels) and $T$ time steps, where each $\mathbf{x}_t = [x_t^1, x_t^2, \dots, x_t^D]^\top \in \mathbb{R}^D$ represents the observations across all variates at time $t$. Given a look-back window of length $L$, the objective is to estimate the next $H$ time steps (the forecast horizon), which yields the forecast of $\mathbf{\hat{X}}_{T+1:T+H} \in \mathbb{R}^{D \times H}$, conditioned on the historical sequence $\mathbf{X}_{T-L+1:T} \in \mathbb{R}^{D \times L}$. 
    The input sequence is processed by a learnable embedding module and projected to the latent dimension \dmodel (detailed in Section~\ref{subsec:embed}) 
    Next, the input embeddings are forwarded to the Transformer backbone. 
    
    A typical Transformer model is constructed by stacking $B$ Transformer blocks, where each block can be represented as follows:
    %
    \begin{align}
        \mathbf{u}^b_t & = \operatorname{Attn} ( \operatorname{Norm} ( \mathbf{h}^{b-1}_t ) ) + \mathbf{h}^{b-1}_t, \label{eq:attn} \\
        \mathbf{h}^b_t & = \operatorname{FFN} ( \operatorname{Norm} ( \mathbf{u}^b_t ) ) + \mathbf{u}^b_t, \label{eq:ffn}
    \end{align}
    %
    where $\operatorname{Attn}(\,\cdot\,)$ denotes the self-attention module, $\operatorname{Norm}(\,\cdot\,)$ are normalization modules, $\operatorname{FFN}(\,\cdot\,)$ denotes the Feed-Forward Network, and $b \in \{0, \dots, B-1\}$ denotes the $b$-th Transformer block~\cite{vaswani2017attention}. Standard Transformers rely on dense computations, where a single shared $\operatorname{FFN}$ processes every token, effectively forcing a ``one-size-fits-all'' transformation. In contrast, sparse $\operatorname{MoE}$ layers enable conditional computation. To introduce sparsity and enable parameter scaling while keeping computational costs, an emerging practice is to replace $\operatorname{FFN}$ modules in a Transformer with $\operatorname{MoE}$ layers. An $\operatorname{MoE}$ layer consists of several sparsely activated expert networks, where each expert is structurally identical to a standard $\operatorname{FFN}$~\cite{shi2024time,dai2024deepseekmoe}. Then, each individual time point can be routed through a gating mechanism to one or more selected experts~\cite{fedus2022switch,gshard} as follows:
    %
    \begin{equation}\label{equ:moe}
        \mathbf{h}^b_t = \operatorname{MoE} ( \operatorname{Norm} ( \mathbf{u}^b_t ) ) + \mathbf{u}^b_t,
    \end{equation}
    %
    with
    %
    \begin{align}
        \operatorname{MoE} & ( \operatorname{Norm} ( \mathbf{u}^b_t ) ) = \sum_{i=1}^{N} ( {g_{i,t} \operatorname{FFN}_{i} ( \operatorname{Norm} ( \mathbf{u}^b_t ) )} ),\label{equ:mixture} \\
        g_{i,t} & = \begin{cases} 
        s_{i,t}, & s_{i,t} \in \operatorname{Topk} (\{ s_{j, t} | 1 \leq j \leq N \}, K), \\
        0, & \text{otherwise}, 
        \end{cases} \label{equ:expert_score} \\
        s_{i,t} & = \operatorname{Softmax}_i ( \mathbf{W}_{i}^{b} ( \operatorname{Norm} ( \mathbf{u}^b_t ) ) ) \label{equ:expert_gate},
    \end{align}
    
    where $N$ represents the number of $\operatorname{FFN}$ experts, $g_{i,t}$ is the gate value for the $i$-th expert, $\operatorname{Topk} (\,\cdot\,, K)$ is the set of $K$ highest affinity scores between the $t$-th time point and all experts, $s_{i,t}$ is the point-to-expert affinity score computed by taking the $\operatorname{Softmax}$ logits from the gate function $\mathbf{W}_{i}^{b} \in \mathbb{R}^{d_{\text{model}} \times N}$, a learnable linear projection~\cite{shazeer2017outrageously}. Therefore, each time point will be forwarded only to $K$ of the $N$ experts, allowing the activated experts to specialize in different time patterns and ensuring computational efficiency~\cite{liu2024moirai,dai2024deepseekmoe}.

    \begin{figure*}[!ht]
        \centering
        \includegraphics[width=0.96\linewidth]{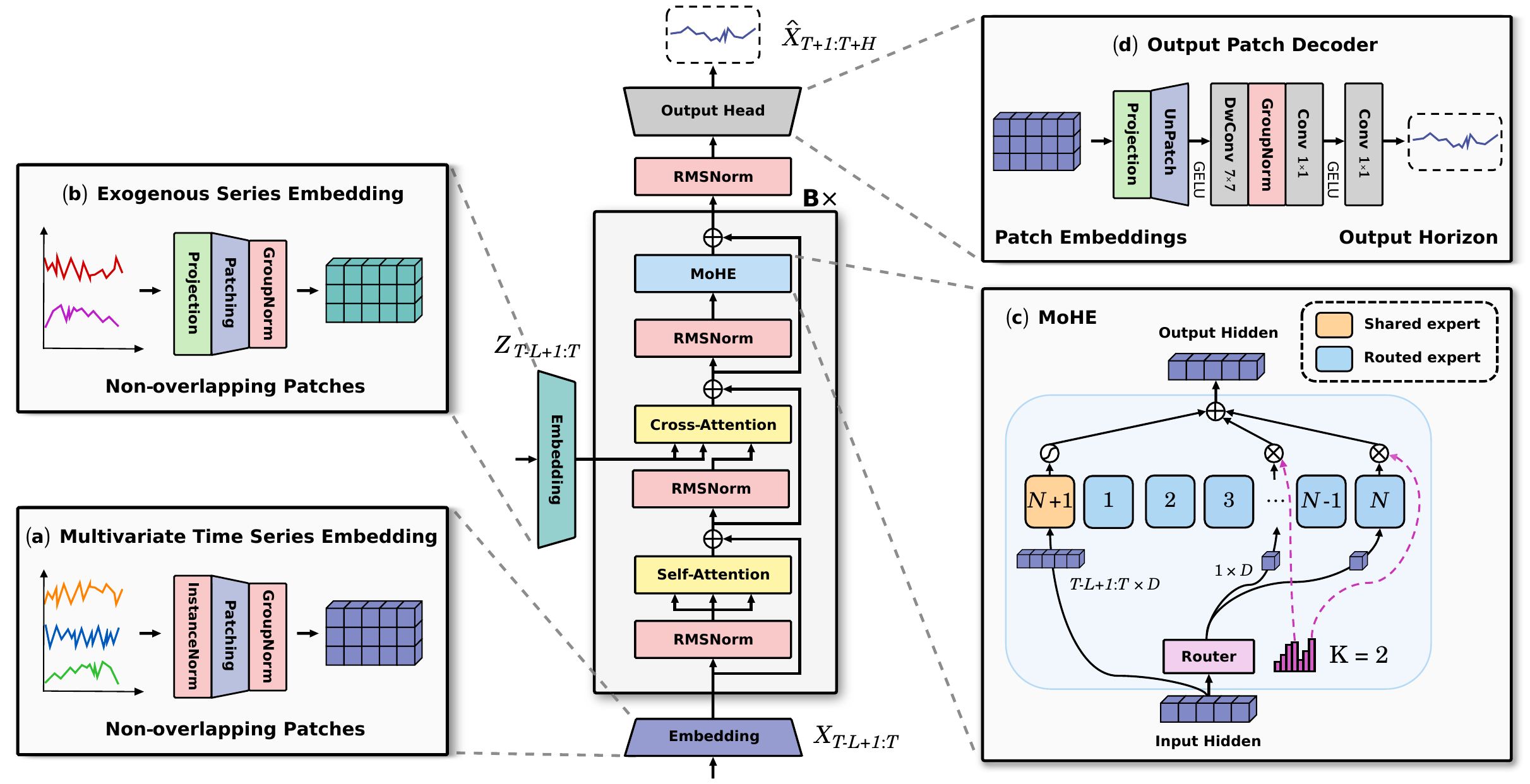}
        \vspace{-2pt}
        \caption{Architecture of \model, an encoder-only transformer for multivariate time-series forecasting. 
        (a) The input embedding module splits time channels into sequences of channel-independent patch embeddings. 
        (b) The exogenous embedding module projects, fuses, and patches covariates with the input series to produce aligned exogenous patch embeddings. 
        These patches are processed through $B$ stacked Transformer blocks; each block is composed of self-attention, cross-attention, and a 
        (c) Mixture-of-Heterogeneous-Experts (MoHE), where a shared depthwise-convolution expert maintains sequence continuity and routed Fourier experts resolve local spectral patterns. 
        (d) The patch decoder head projects final embeddings to forecasting horizons.}
        \label{fig:model_arch}
    \end{figure*}

\subsection{\model Architecture}
\label{subsec:overview}

    Our proposed \model, illustrated in \Figref{fig:model_arch}, adopts an encoder-only Transformer backbone that leverages recent advances in stability optimizations from large-scale language models~\cite{grattafiori2024llama,dai2024deepseekmoe}. 
    Specifically, we employ $\operatorname{RMSNorm}$~\cite{zhang2019root} in the $\operatorname{Norm}$ modules to normalize inputs to each Transformer sub-layer, enhancing training stability \cite{xiong2020layer}. 
    In forecasting, the relative distance between time steps (e.g., ``24 hours ago'') carries more predictive signal than absolute indices \cite{erturk2025beyond}. Consequently, we replace standard additive positional encodings with Rotary Position Embeddings (RoPE) \cite{su2024roformer}. By injecting position information directly into the attention query-key products, RoPE improves the model's ability to extrapolate to unseen future horizons. 
    Finally, to structurally decouple the modeling of global trends and local periodicities, we replace dense $\operatorname{FFNs}$ with our $\operatorname{MoHE}$ layers. This design enables conditional computation by dynamically routing patches to the operators best suited to their signal characteristics.
    
\subsection{Input Embedding}
\label{subsec:embed}

    To mitigate distribution shift issues in non-stationary time series, we apply instance normalization to the input sequence~\cite{liu2022non}, which normalizes each input variate by its instance mean and variance before patching and denormalizes the output predictions to restore the original scale. 
    Standard Transformers process time series sequences as time-point tokens, with quadratic complexity in the look-back length $L$. To address this limitation, patching techniques segment the input sequence into subseries~\cite{nie2022time}, treating each patch as a semantic token~\cite{liu2024moirai}. Specifically, for a patch length $P$, the look-back window $\mathbf{X}_{T-L+1:T}$ is embedded into $S = \lceil L / P \rceil$ non-overlapping patches $\mathbf{P} = \{ \mathbf{p}_1, \mathbf{p}_2, \dots, \mathbf{p}_S \}$, $\mathbf{p}_i \in \mathbb{R}^{D \times P}$, effectively decreasing the quadratic attention complexity to $O(S^2)$ while aggregating high-frequency local noise into robust feature vectors. Our patch embedding process is illustrated in \Figref{fig:model_arch}a and defined as:
    %
    \begin{equation}
        \begin{aligned}
            & \{\mathbf{p}_1, \mathbf{p}_2, \dots, \mathbf{p}_S\} = \operatorname{Patchify} (\mathbf{x}), \\
            & \mathbf{h}^0_{p} = \operatorname{PatchEmbed} ( \operatorname{GNorm} ( \mathbf{p} ) ), \\
        \end{aligned}
        \label{embedding}
    \end{equation}
    %
    \hspace{-2pt}where \(\mathbf{h}^0_i \in \mathbb{R}^{d_{\text{model}}}\) is the embedded representation of patch \(\mathbf{p}_i\) and $\operatorname{GNorm}$ applies a single-group normalization to the embedding dimension~\cite{wu2018group}. Furthermore, the channel independence approach~\cite{nie2022time} is used to process each variate of a multivariate input as a univariate series, allowing \model to operate in any-variate forecasting tasks.

\subsection{Self-Attention}
\label{subsec:sattn}

    To capture temporal dependencies across patch embeddings, we implement multi-head self-attention (see Equation~\ref{eq:attn}) optimized for efficiency and extrapolation. To efficiently handle extended look-back windows (L), we employ FlashAttention-2~\cite{dao2023flashattention}. This implementation reformulates the scaled dot-product to reduce memory access costs from quadratic to linear with respect to sequence length, enabling rapid training on long sequences.
    We further reduce the inference memory footprint by implementing grouped-query attention (GQA)~\cite{ainslie2023gqa}.
    By sharing a single key-value head across multiple query heads, GQA maintains the expressivity of multi-head attention while significantly reducing the memory overhead needed for autoregressive rollout. 
    Following \citet{shi2024time}, we adopt a bias-free architecture for all layers except the QKV projections. Retaining QKV biases has been shown to preserve length extrapolation capabilities by maintaining shift-invariance in the attention scores~\cite{chowdhery2023palm}.
    In addition, we apply channel-independent attention per variate~\cite{nie2022time}. This mechanism forces the model to learn universal temporal dynamics (e.g., seasonality) that generalize across different variates, preventing overfitting to spurious cross-variate correlations. 
    %
    %
    Details about our implementation of multi-head attention are in Appendix~\ref{apdx:technical}.

\subsection{Multimodal Cross-Attention}
\label{subsec:cattn}

    Let $\mathbf{Z}_{T-L+1:T+H} \in \mathbb{R}^{C \times (L+H)}$ denote an additional sequence with $C$ covariate dimensions, such as calendar indicators or weather metrics, assumed to be known over the forecast horizon~\cite{wang2024timexer}. We design a multimodal embedding module that incorporates exogenous covariate information into endogenous time series. The module operates in two stages. First, we use linear layers to project both endogenous and covariate sequences to the model dimension \dmodel:
    %
    \begin{equation}
        \begin{aligned}
            \mathbf{W}^X_t &= \operatorname{Linear}_X(\mathbf{x}_t), \quad \mathbf{x}_t \in \mathbb{R}^D, \quad \mathbf{W}^X_t \in \mathbb{R}^{d_{\text{model}}}, \\
            \mathbf{W}^Z_t &= \operatorname{Linear}_Z(\mathbf{z}_t), \quad \mathbf{z}_t \in \mathbb{R}^C, \quad \mathbf{W}^Z_t \in \mathbb{R}^{d_{\text{model}}},
        \end{aligned}
        \label{eq:multimodal_embedding}
    \end{equation}
    %
    \hspace{-3pt}These projections are fused via concatenation and a subsequent projection layer: $\mathbf{w}^M_t = \operatorname{Linear}_{\text{ fuse}}([\mathbf{w}^X_t; \mathbf{w}^Z_t])$, creating a covariate-enriched latent representation aligned with the endogenous dimension $D$. 
    Concatenation followed by linear projection outperforms alternatives such as element-wise addition or direct concatenation~\cite{bao2023all}.
    As illustrated in \Figref{fig:model_arch}b, the fused sequence $\mathbf{W}^{M}_{t} \in \mathbb{R}^{D \times L}$ is then patched and embedded into $S = \lceil L / P \rceil$ non-overlapping tokens, serving as keys and values for the cross-attention mechanism. 
    %
    %
    By adopting this approach, we ensure that exogenous information is aggregated into endogenous data regardless of the difference between $D$ and $C$.

    In the cross-attention module, the self-attention output (Section~\ref{subsec:sattn}) serves as the query, while the multimodal patch embeddings feed the key and value projections \cite{wang2024timexer}. This design allows the model to dynamically retrieve external context such as ``holiday effects'' or ``weather spikes'', conditional on the current state. 
    Cross-attention enables a Transformer model to combine information from different modalities, acting as a connector between representations from one modality by attending to another.
    Our cross-attention module computes multi-head attention with optimizations that mirror those of the self-attention module (Section~\ref{subsec:sattn}), specifically FlashAttention \cite{dao2023flashattention} for efficient scaled dot-product computation and GQA \cite{ainslie2023gqa} to reduce memory overhead by clustering queries and sharing key-value pairs. The above process can be formalized as
    %
    \begin{equation}\label{eq:cattn}
        \mathbf{v}^b_{p} = \operatorname{CrossAttn} ( \operatorname{RMSNorm} ( \mathbf{u}^b_{p} ), \mathbf{h}^{}_{m} ) + \mathbf{u}^b_{p}.
    \end{equation}
    %
    \hspace{-6pt}Channel independence is maintained through variate-wise processing, ensuring alignment with the endogenous pipeline and scalable covariate integration. Since covariates $\mathbf{Z}$ are known for the future horizon \(t = T+1, \dots, T+H\), this fusion process is repeated during the autoregressive rollout, allowing the model to anticipate future external temporal information before it appears in the series.

\subsection{Mixture-of-Heterogeneous-Experts (MoHE)}
\label{subsec:mohe}

    To enhance specialization and scalability in our Transformer forecaster, we follow recent advances in time series models and replace dense $\operatorname{FFNs}$ with sparse $\operatorname{MoE}$ layers~\cite{shi2024time}. 
    However, standard $\operatorname{MoEs}$ are based on homogeneous MLP experts, assuming that all tokens require the same processing logic. We argue that this approach is suboptimal for time series, which are a superposition of global trends and local frequencies. We therefore propose the Mixture-of-Heterogeneous-Experts ($\operatorname{MoHE}$) design based on a structural labor partitioning: assigning sequence-continuity modeling to a shared expert and local-frequency analysis to routed experts.
    Combining routed experts working at the unit level with a shared expert, which is always activated and operates at the sequence level to capture and consolidate common knowledge, improves specialization robustness~\cite{dai2024deepseekmoe,shi2024time}.

    Each $\operatorname{MoHE}$ expert, shared or routed, comprises two layers that project time patches from the model dimension \dmodel to an intermediate dimension \dff and back to \dmodel, with a factor of \dff~$= 2 \times$\dmodel, GELU activation~\cite{hendrycks2016gaussian}, and dropout~\cite{srivastava2014dropout}. 
    The shared expert is a depthwise separable convolution ($\operatorname{DwConvFFN}$) that acts as a time-domain expert: it slides along the sequence dimension to capture continuous trends and maintain temporal coherence across patches. Conversely, the routed experts are Fourier-based networks ($\operatorname{FA-FFN}$) acting as frequency-domain experts. By operating in the spectral domain, $\operatorname{FA-FFNs}$ isolate high-frequency periodicities within individual patches -- a task where standard MLPs often struggle due to spectral bias~\cite{dong2024fan}. 
    Specifically, a $\operatorname{FA-FFN}$ replaces general-purpose linear layers with $\operatorname{FAN}$ modules~\cite{dong2024fan}, which are designed to leverage the strengths of Fourier series transformations to model periodic patterns from time series signals.

    Therefore, a $\operatorname{MoHE}$ layer combines one $\operatorname{DwConvFFN}$ shared expert and \(N\) $\operatorname{FA-FFN}$ routed experts:
    %
    \begin{equation}
        \begin{aligned}
            & \mathbf{\bar{v}}^b_{p} = \operatorname{RMSNorm} ( \mathbf{v}^b_{p} ), \\
            \operatorname{MoHE} ( \mathbf{\bar{v}}^b_{p} ) & = g_{N+1,p} \operatorname{DwConvFFN}_{N+1} ( \mathbf{\bar{v}}^b_{p} ) \\
            & + \sum_{i=1}^{N} ( {g_{i,p} \operatorname{FA-FFN}_{i} ( \mathbf{\bar{v}}^b_{p} )} ), \label{equ:mohe}
        \end{aligned}
    \end{equation}
    %
    where $g_{N + 1,p}$ denotes a $\operatorname{Sigmoid}$ function gate, modulating the shared expert contribution~\cite{shi2024time}, and $g_{i,p}$ represents the router gate value defined in Equation~\ref{equ:expert_score} (dotted lines in \Figref{fig:model_arch}c).
    
    The heterogeneity introduced by the $\operatorname{MoHE}$ design draws inspiration from the trend-seasonality decomposition often used in statistical forecasting \cite{wu2021autoformer}. However, instead of hard-coding the decomposition, $\operatorname{MoHE}$ learns to dynamically route signal components: directing transient noise and short-term periodicity to Fourier experts while reserving the shared convolutional path for persistent sequence-level trends.
    Sparse architectures dynamically activate different experts to handle heterogeneous data patterns, with each expert specializing in learning different knowledge~\cite{shi2024time}.
    This $\operatorname{MoHE}$ design improves specialization over homogeneous $\operatorname{MoEs}$, thereby enhancing generalization and forecasting accuracy for heterogeneous temporal patterns.

\subsection{Output Patch Decoder}
\label{subsec:output}

    To map the Transformer's output from patches to forecast time points, most of the state-of-the-art time series models have implemented linear-based projection heads~\cite{nie2022time,liu2023itransformer,wang2024timexer,shi2024time}. 
    Standard linear heads flatten the patch embeddings, destroying the local temporal structure preserved by the encoder. Furthermore, the parameter count of a linear head scales as $O(L \times D)$, leading to parameter explosion and overfitting when the number of variates $D$ increases.
    We replace linear projections with a convolutional patch decoder module (see \Figref{fig:model_arch}d) that projects the latent dimension \dmodel to time points using a lightweight sequence of convolutions \cite{convnet}. This design imposes a locality inductive bias, ensuring that the output generation relies on the semantic vectors of each patch rather than a global dense matrix, stabilizing training.

    Each variate is processed independently by our output patch decoder, maintaining channel independence to handle any-variate forecasting. 
    With this convolutional patch decoder, we provide a lightweight module that mitigates the instability of heavy linear heads, helping \model to achieve superior accuracy in forecasting tasks. 
    Refer to Appendix~\ref{apdx:technical} for technical details.

\subsection{Loss Functions}
\label{subsec:loss}

    Time series data often contain transient outliers and extreme spikes that can destabilize training \cite{wen2019robusttrend}, especially for sparse models where gradients are routed to specific experts \cite{fedus2022switch}. Standard MSE amplifies these outliers quadratically. Therefore, we adopt the Huber loss \cite{huber1992robust} as our main prediction loss ($\mathcal{L}_{\text{pred}}$), which behaves quadratically for small errors and linearly for large errors, providing a robust gradient signal that prevents expert collapse due to outlier-driven gradients.

    However, focusing only on prediction error optimization often leads to stability and convergence challenges due to load imbalance issues when using $\operatorname{MoE}$ architectures. Specifically, the sparse gating mechanism introduces the risk of routing collapse~\cite{shazeer2017outrageously}, in which the model converges to a trivial state by selecting a single expert, limiting the opportunities for other experts to receive sufficient training~\cite{dai2024deepseekmoe}. To prevent routing collapse, we impose an auxiliary load balancing loss ($\mathcal{L}_{\text{aux}}$) to balance expert utilization~\cite{gshard,fedus2022switch,shi2024time}. This loss minimizes the coefficient of variation of the expert assignment probabilities, ensuring a more uniform distribution of tokens across the experts' pool.
    We detail the prediction and balance losses in Appendix~\ref{apdx:technical}.

\subsection{Training Objective and Forecasting}
\label{subsec:objective}

    The training objective combines the prediction loss with the auxiliary balance loss to compose the final loss, ensuring both accuracy and expert balance: 
    %
    \begin{equation}
        \mathcal{L} = \mathcal{L}_{\text{pred}} \left(\mathbf{{X}}_{T+1:T+H_o}, \hat{\mathbf{X}}_{T+1:T+H_o} \right) + \alpha \mathcal{L}_{\text{aux}},
    \end{equation}
    %
    \hspace{-1pt}where $H_o$ is the length of the predicted future time steps and $\alpha$ is the expert balance factor. The patch length is uniformly set as $P$ from the input embedding to the output patch decoder, with \model supporting flexible output resolutions $H_o$.
    
    Time series models have been trained with different look-back lengths, ranging from small~\cite{wang2024timexer} to large values~\cite{shi2024time}. Very small look-back windows lack contextual information critical for long-term forecasting, while very long windows can introduce undesirable noise. 
    We standardize the look-back window to $L = 672$ (approximately 4 weeks of hourly data). This duration balances the need to capture monthly seasonality with the computational cost of attention, avoiding the diminishing returns often observed with ultra-long contexts in noisy real-world data~\cite{liu2024timer}. 
    \model is trained as a one-for-all forecaster~\cite{liu2024timerxl}. During inference, we employ autoregressive rolling forecasting: the model predicts a fixed chunk ($H_o$ next steps), which is appended to the input buffer to predict the subsequent chunk. This approach allows a single trained model to generate arbitrary horizons (e.g., $\{96, 192, 336, 720\}$) without retraining for any specific length.

\subsection{Model Settings and Training Details}
\label{subsec:settings}

    \model and all experiments are implemented in PyTorch~\cite{paszke2019pytorch}. Trainings are performed on a single NVIDIA A100 (80GB) GPU. To maximize throughput without sacrificing numerical stability, we utilized TensorFloat-32 (TF32) precision, which provides the dynamic range of FP32 with the matrix-multiplication speed of FP16.
    We experiment with the number of Transformer blocks searched from $B\in\{4, 6, 8\}$, the dimension of representations \dmodel from $\{64, 128, 256, 384\}$, the patch length as $P \in \{8, 12, 16\}$, and the output resolution as $H_o \in \{16, 24, 32\}$ time points.
    Following the results achieved by~\cite{dai2024deepseekmoe,liu2024moirai,shi2024time}, we set $8$ as the number of experts per $\operatorname{MoHE}$ layer with $K = 2$, which is an optimal choice to balance performance and computational efficiency~\cite{shi2024time}.

    For optimization, we apply the fused implementation of AdamW~\cite{loshchilov2017decoupled} with a maximum learning rate of $3.2 \times 10^{-3}$, $\text{weight\_decay}=1 \times 10^{-4}$, $\beta_1=0.9$, and $\beta_2=0.95$~\cite{chen2020generative}. 
    Furthermore, we use a Cosine Annealing scheduler with a linear warmup~\cite{loshchilov2016sgdr} for the first 10\% training steps, and a decay to a minimum learning rate of $1.2 \times 10^{-4}$ and early stopping with $\text{patience}=5$.
    Such an aggressive warmup is essential to stabilize the router's initial random assignment before the experts begin specializing.
    Following \cite{shi2024time}, we set the Huber loss $\delta$ to $2.0$ and the auxiliary loss factor $\alpha$ to $0.02$. 
    Training epochs are searched from $10$ to $30$. We do not use data augmentations. 
    Refer to Appendix~\ref{apdx:settings} for more optimization details.

    \begin{table*}[!htb]
        \centering
        \caption{Results of long-term multivariate forecasting experiments. Full-shot results are obtained from~\cite{liu2023itransformer,wang2024timexer,han2024softs}. {\bestres{Bold red}}: the best, \secondres{underlined blue}: the second best. Results are averaged from all prediction horizons $H = \{96, 192, 336, 720\}$. $1^{\text{st}}$ Count represents the number of wins achieved by a model.}
        \label{tab:ltsf_avg_full}
        \resizebox{\linewidth}{!}{
        \renewcommand{\tabcolsep}{3pt}
        \begin{tabular}{c|cc|cc|cc|cc|cc|cc|cc|cc|cc|cc|cc}
        \toprule
            \multicolumn{1}{c}{\scalebox{1}{\textbf{Models}}} & \multicolumn{2}{c}{\textbf{{\model}}} & \multicolumn{2}{c}{\textbf{SOFTS}} & \multicolumn{2}{c}{\textbf{TimeXer}} & \multicolumn{2}{c}{\textbf{iTransformer}} & \multicolumn{2}{c}{\textbf{TimeMixer}} &  \multicolumn{2}{c}{\textbf{TimesNet}} & \multicolumn{2}{c}{\textbf{PatchTST}} & \multicolumn{2}{c}{\textbf{Crossformer}} & \multicolumn{2}{c}{\textbf{TiDE}} & \multicolumn{2}{c}{\textbf{DLinear}} & \multicolumn{2}{c}{\textbf{FEDformer}} \\
    
            \cmidrule(lr){2-3} \cmidrule(lr){4-5}\cmidrule(lr){6-7} \cmidrule(lr){8-9}\cmidrule(lr){10-11}\cmidrule(lr){12-13}\cmidrule(lr){14-15}\cmidrule(lr){16-17}\cmidrule(lr){18-19} \cmidrule(lr){20-21} \cmidrule(lr){22-23}

            \multicolumn{1}{c}{\scalebox{1}{\textbf{Metrics $\downarrow$}}} & \textbf{MSE} & \textbf{MAE} & \textbf{MSE} & \textbf{MAE} & \textbf{MSE} & \textbf{MAE} & \textbf{MSE} & \textbf{MAE} & \textbf{MSE} & \textbf{MAE} & \textbf{MSE} & \textbf{MAE} & \textbf{MSE} & \textbf{MAE} & \textbf{MSE} & \textbf{MAE} & \textbf{MSE} & \textbf{MAE} & \textbf{MSE} & \textbf{MAE} & \textbf{MSE} & \textbf{MAE} \\
        \midrule
            {ETTh1} & \bestres{0.383} & \bestres{0.412} & 0.449 & 0.442 & \secondres{0.437} & \secondres{0.437} & 0.454 & 0.447 & 0.448 & 0.442 & 0.454 & 0.450 & 0.469 & 0.454 & 0.529 & 0.522 & 0.540 & 0.507 & 0.455 & 0.451 & 0.440 & 0.459 \\
        \midrule
            {ETTh2} & \bestres{0.348} & \bestres{0.382} & 0.373 & 0.400 & 0.367 & 0.396 & 0.383 & 0.406 & \secondres{0.364} & \secondres{0.395} & 0.414 & 0.496 & 0.387 & 0.407 & 0.942 & 0.683 & 0.611 & 0.549 & 0.558 & 0.515 & 0.436 & 0.449 \\
        \midrule
            {ETTm1} & \bestres{0.333} & \bestres{0.367} & 0.393 & 0.403 & 0.382 & 0.397 & 0.407 & 0.410 & \secondres{0.381} & \secondres{0.395} & 0.400 & 0.405 & 0.387 & 0.400 & 0.513 & 0.495 & 0.419 & 0.419 & 0.403 & 0.406 & 0.448 & 0.452 \\          
        \midrule
            {ETTm2} & \bestres{0.256} & \bestres{0.310} & 0.287 & 0.330 & \secondres{0.274} & \secondres{0.322} & 0.288 & 0.332 & 0.275 & 0.323 & 0.291 & 0.332 & 0.281 & 0.326 & 0.757 & 0.610 & 0.358 & 0.403 & 0.350 & 0.400 & 0.304 & 0.349 \\
        \midrule
            {Weather} & \bestres{0.216} & \bestres{0.249} & 0.255 & 0.278 & 0.241 & \secondres{0.271} & 0.258 & 0.278 & \secondres{0.240} & \secondres{0.271} & 0.259 & 0.286 & 0.259 & 0.281 & 0.258 & 0.315 & 0.270 & 0.320 & 0.265 & 0.316 & 0.308 & 0.360 \\ 
        \midrule
            {ECL} & \bestres{0.158} & \bestres{0.251} & 0.174 & \secondres{0.264} & \secondres{0.171} & 0.270 & 0.178 & 0.270 & 0.182 & 0.272 & 0.192 & 0.295 & 0.205 & 0.290 & 0.244 & 0.334 & 0.251 & 0.344 & 0.212 & 0.300 & 0.214 & 0.327 \\
        \midrule
            {Traffic} & \bestres{0.388} & \bestres{0.256} & \secondres{0.409} & \secondres{0.267} & 0.466 & 0.287 & 0.428 & 0.282 & 0.484 & 0.297 & 0.620 & 0.336 & 0.481 & 0.304 & 0.550 & 0.304 & 0.760 & 0.473 & 0.625 & 0.383 & 0.609 & 0.376 \\
        \midrule
            {\textbf{Average}} & \bestres{0.297} & \bestres{0.318} & \secondres{0.334} & 0.341 & \secondres{0.334} & \secondres{0.340} & 0.342 & 0.346 & 0.339 & 0.342 & 0.376 & 0.371 & 0.353 & 0.352 & 0.542 & 0.466 & 0.458 & 0.431 & 0.410 & 0.396 & 0.394 & 0.396 \\
        \midrule
            \rowcolor{tabhighlight}{\textbf{{$1^{\text{st}}$ Count}}} & \multicolumn{2}{c}{16} & \multicolumn{2}{c}{0} & \multicolumn{2}{c}{0} & \multicolumn{2}{c}{0} & \multicolumn{2}{c}{0} & \multicolumn{2}{c}{0} & \multicolumn{2}{c}{0} & \multicolumn{2}{c}{0} &  \multicolumn{2}{c}{0} & \multicolumn{2}{c}{0} & \multicolumn{2}{c}{0} \\
        \bottomrule
        \end{tabular}%
        }
    \end{table*}%

\section{Main Results}
\label{sec:results}
    
    We conduct extensive experiments to evaluate the performance and efficiency of \model in long-term multivariate forecasting tasks. 
    Our experiments include $7$ benchmark datasets that cover a wide variety of real-world domains with different temporal resolutions and number of variables, as well as $15$ baseline models representing the state-of-the-art in long-term forecasting. 
    A detailed summary of each benchmark data is provided in Appendix~\ref{apdx:datasets} while the baseline models are presented in Appendix~\ref{apdx:baselines}. 
    As evaluation metrics, we adopt the mean squared error (MSE) and the mean absolute error (MAE), detailed in Appendix~\ref{apdx:metrics}.
    
    To ensure fair comparisons, we follow the data processing and train-validation-test split protocol defined by~\cite{wu2023timesnet}, where the train, validation, and test datasets are split chronologically to prevent data leakage. 
    Each benchmark is performed over four long-term prediction horizons, which are $H \in \{96, 192, 336, 720\}$.
    Furthermore, we feed the cross-attention modules of \model only with calendar information derived from the timestamp of each instance, splitting the calendar components and scaling them to continuous linear frequency values~\cite{gluonts_jmlr}.

\subsection{Multivariate Time Series Forecasting}
\label{subsec:forecasting}

    \Tabref{tab:ltsf_avg_full} shows long-term multivariate forecasting results. 
    \model establishes a new state-of-the-art, surpassing the strongest baseline (TimeXer, SOFTS) on all datasets. In particular, it dominates on datasets with strong seasonality (ETTh1, ETTm2), validating the $\operatorname{MoHE}$ architecture's ability to decouple periodic patterns.
    When analyzing each average performance over the horizons $\{96, 192, 336, 720\}$, \model reduces the average MSE by $12.3\%$ relative to TimeXer on the ETTh1 dataset, by $10\%$ relative to TimeMixer on Weather, and by $5.1\%$ relative to SOFTS on the Traffic data, demonstrating the efficacy of its MoHE architecture and support for covariate injection.
    The complete results and comparisons with time series foundation models are included in Appendix~\ref{apdx:results}.

\subsection{Ablation Study}
\label{subsec:ablation}

    \begin{table*}[!htb]
        \caption{Ablation study with different Transformer types. The best results are in \textbf{bold}.}
        \label{tab:abl_encoder_decoder}
        \centering
        \begin{small}
        \resizebox{0.91\linewidth}{!}{
        \renewcommand{\multirowsetup}{\centering}
        \setlength{\tabcolsep}{6pt}
        \begin{tabular}{K{2.6cm}|cc|cc|cc|cc|cc|cc|cc}
        \toprule
            \multicolumn{1}{c}{\textbf{Dataset}} & \multicolumn{2}{c}{\textbf{ETTh1}} & \multicolumn{2}{c}{\textbf{ETTh2}} & \multicolumn{2}{c}{\textbf{ETTm1}} & \multicolumn{2}{c}{\textbf{ETTm2}} & \multicolumn{2}{c}{\textbf{Weather}} & \multicolumn{2}{c}{\textbf{ECL}} & \multicolumn{2}{c}{\textbf{Traffic}} \\
            \cmidrule(lr){2-3} \cmidrule(lr){4-5} \cmidrule(lr){6-7} \cmidrule(lr){8-9} \cmidrule(lr){10-11} \cmidrule(lr){12-13} \cmidrule(lr){14-15}
            \multicolumn{1}{c}{\textbf{Metrics (Avg.) $\downarrow$}} & \textbf{MSE} & \textbf{MAE} & \textbf{MSE} & \textbf{MAE} & \textbf{MSE} & \textbf{MAE} & \textbf{MSE} & \textbf{MAE} & \textbf{MSE} & \textbf{MAE} & \textbf{MSE} & \textbf{MAE} & \textbf{MSE} & \textbf{MAE} \\
        \midrule
            \rowcolor{tabhighlight}Encoder & \textbf{0.383} & \textbf{0.415} & \textbf{0.362} & \textbf{0.389} & \textbf{0.338} & \textbf{0.366} & 0.270 & \textbf{0.313} & \textbf{0.244} & \textbf{0.265} & \textbf{0.164} & \textbf{0.254} & \textbf{0.406} & \textbf{0.276} \\
            Decoder & 0.417 & 0.430 & 0.371 & 0.395 & 0.354 & 0.375 & \textbf{0.269} & 0.315 & 0.255 & 0.270 & 0.167 & 0.262 & 0.449 & 0.297 \\
        \bottomrule
        \end{tabular}
        }
        \end{small}
    \end{table*}

    \begin{table*}[!htb]
        \caption{Ablation study with different expert architectures. The best results are in \textbf{bold} and the second best are \underline{underlined}.}
        \label{tab:abl_mohe}
        \centering
        \begin{small}
        \resizebox{0.91\linewidth}{!}{
        \renewcommand{\multirowsetup}{\centering}
        \setlength{\tabcolsep}{6pt}
        \begin{tabular}{K{2.6cm}|cc|cc|cc|cc|cc|cc|cc}
        \toprule
            \multicolumn{1}{c}{\textbf{Dataset}} & \multicolumn{2}{c}{\textbf{ETTh1}} & \multicolumn{2}{c}{\textbf{ETTh2}} & \multicolumn{2}{c}{\textbf{ETTm1}} & \multicolumn{2}{c}{\textbf{ETTm2}} & \multicolumn{2}{c}{\textbf{Weather}} & \multicolumn{2}{c}{\textbf{ECL}} & \multicolumn{2}{c}{\textbf{Traffic}} \\
            \cmidrule(lr){2-3} \cmidrule(lr){4-5} \cmidrule(lr){6-7} \cmidrule(lr){8-9} \cmidrule(lr){10-11} \cmidrule(lr){12-13} \cmidrule(lr){14-15}
            \multicolumn{1}{c}{\textbf{Metrics (Avg.) $\downarrow$}} & \textbf{MSE} & \textbf{MAE} & \textbf{MSE} & \textbf{MAE} & \textbf{MSE} & \textbf{MAE} & \textbf{MSE} & \textbf{MAE} & \textbf{MSE} & \textbf{MAE} & \textbf{MSE} & \textbf{MAE} & \textbf{MSE} & \textbf{MAE} \\
        \midrule
            \rowcolor{tabhighlight}\model & \textbf{0.383} & \textbf{0.415} & 0.362 & \textbf{0.389} & \textbf{0.338} & \textbf{0.366} & 0.270 & \underline{0.313} & 0.244 & \underline{0.265} & \textbf{0.164} & \textbf{0.254} & \textbf{0.406} & \textbf{0.276} \\
            $\operatorname{MLP}$ & 0.399 & \underline{0.420} & \underline{0.361} & 0.392 & \underline{0.348} & \underline{0.373} & \underline{0.264} & 0.314 & 0.262 & 0.272 & 0.267 & 0.318 & 0.498 & 0.331 \\
            $\operatorname{FA}$ & \underline{0.391} & 0.421 & 0.364 & \underline{0.390} & 0.352 & 0.377 & 0.268 & 0.314 & \textbf{0.230} & \textbf{0.262} & \underline{0.174} & 0.269 & 0.428 & 0.290 \\
            $\operatorname{Conv + MLP}$ & 0.398 & \underline{0.420} & \textbf{0.357} & \textbf{0.389} & 0.353 & 0.378 & \textbf{0.260} & \textbf{0.312} & \underline{0.234} & \underline{0.265} & 0.197 & 0.293 & 0.435 & \underline{0.284} \\
            $\operatorname{Conv + FA}$ & 0.402 & 0.425 & 0.367 & 0.395 & 0.356 & 0.376 & 0.269 & \textbf{0.312} & 0.237 & 0.266 & \textbf{0.164} & \underline{0.255} & \underline{0.413} & \textbf{0.276} \\
            $\operatorname{DwConv + MLP}$ & 0.401 & 0.424 & 0.372 & 0.404 & 0.368 & 0.375 & 0.279 & 0.325 & 0.368 & 0.283 & 0.239 & 0.325 & 0.420 & \underline{0.284} \\
        \bottomrule
        \end{tabular}
        }
        \end{small}
    \end{table*}

    \begin{table*}[!htb]
        \caption{Ablation study using different normalization layers, projection head, and with no covariate injection. The best results are in \textbf{bold} and the second best are \underline{underlined}.}
        \label{tab:abl_norm}
        \centering
        \begin{small}
        \resizebox{0.91\linewidth}{!}{
        \renewcommand{\multirowsetup}{\centering}
        \setlength{\tabcolsep}{6pt}
        \begin{tabular}{K{2.65cm}|cc|cc|cc|cc|cc|cc|cc}
        \toprule
            \multicolumn{1}{c}{\textbf{Dataset}} & \multicolumn{2}{c}{\textbf{ETTh1}} & \multicolumn{2}{c}{\textbf{ETTh2}} & \multicolumn{2}{c}{\textbf{ETTm1}} & \multicolumn{2}{c}{\textbf{ETTm2}} & \multicolumn{2}{c}{\textbf{Weather}} & \multicolumn{2}{c}{\textbf{ECL}} & \multicolumn{2}{c}{\textbf{Traffic}} \\
            \cmidrule(lr){2-3} \cmidrule(lr){4-5} \cmidrule(lr){6-7} \cmidrule(lr){8-9} \cmidrule(lr){10-11} \cmidrule(lr){12-13} \cmidrule(lr){14-15}
            \multicolumn{1}{c}{\textbf{Metrics (Avg.) $\downarrow$}} & \textbf{MSE} & \textbf{MAE} & \textbf{MSE} & \textbf{MAE} & \textbf{MSE} & \textbf{MAE} & \textbf{MSE} & \textbf{MAE} & \textbf{MSE} & \textbf{MAE} & \textbf{MSE} & \textbf{MAE} & \textbf{MSE} & \textbf{MAE} \\
        \midrule
            \rowcolor{tabhighlight}\model & \textbf{0.383} & \textbf{0.415} & \underline{0.362} & \textbf{0.389} & \textbf{0.338} & \textbf{0.366} & \underline{0.270} & \textbf{0.313} & 0.244 & \underline{0.265} & \textbf{0.164} & \textbf{0.254} & \textbf{0.406} & 0.276 \\
            $\operatorname{LayerNorm}$ & 0.425 & 0.437 & 0.458 & 0.442 & 0.362 & 0.379 & 0.280 & 0.326 & 0.281 & 0.284 & \textbf{0.164} & \underline{0.256} & 0.413 & \textbf{0.270} \\
            $\operatorname{RMSNorm}$ & 0.436 & 0.444 & 0.439 & 0.436 & 0.358 & 0.377 & 0.278 & 0.322 & \textbf{0.229} & \textbf{0.259} & 0.188 & 0.274 & 0.430 & 0.288 \\
            MLP-based head & 0.457 & 0.451 & \textbf{0.354} & \underline{0.392} & \underline{0.351} & 0.380 & \textbf{0.259} & \underline{0.314} & \underline{0.237} & 0.272 & 0.177 & 0.276 & 0.418 & 0.301 \\
            w/o exogenous covs. & \underline{0.418} & \underline{0.429} & 0.381 & 0.398 & 0.357 & \underline{0.375} & 0.283 & 0.323 & 0.247 & 0.269 & \underline{0.174} & 0.264 & \underline{0.409} & \underline{0.273} \\
        \bottomrule
        \end{tabular}
        }
        \end{small}
    \end{table*}

    We validate the effectiveness of our designs in \model by conducting detailed ablation studies on key architectural components using experimental benchmarks, including the Transformer backbone type, various compositions of our $\operatorname{MoHE}$ approach, the incorporation of exogenous covariates, normalization, and the final projection head. We fix \dmodel $=128$ (see Section~\ref{subsec:settings}) and limit the maximum number of training epochs to $20$ in each ablation experiment to save computational time. All results are averaged over the horizons $H \in \{96, 192, 336, 720\}$, with lower MSE or MAE values indicating better performance.

    \vspace{1pt}\textbf{Transformer Backbone Type.} We compare \model with encoder-only and decoder-only backbones. 
    \Tabref{tab:abl_encoder_decoder} shows that the encoder-only design outperforms the decoder-only variant, achieving average MSE reductions of $8.1\%$ on ETTh1 and $9.6\%$ on Traffic. 
    Additionally, the encoder-only approach is more flexible in supporting longer output resolutions, which can help mitigate the accumulation of autoregressive errors in long-horizon forecasting.

    \vspace{1pt}\textbf{MoHE Configurations.} Once the encoder-only architecture is established as our default backbone, we ablate different expert compositions for the $\operatorname{MoHE}$ layers: all $\operatorname{MLP-FFNs}$ (i.e., a standard $\operatorname{MoE}$ as defined in Equation~\ref{equ:mixture}), all $\operatorname{FA-FFNs}$, mixing a $\operatorname{DwConvFFN}$ with $\operatorname{MLP-FFNs}$, and mixing a simple $\operatorname{ConvFFN}$, composed of two pointwise convolution layers, with $\operatorname{MLP-FFNs}$ or $\operatorname{FA-FFNs}$. 
    \Tabref{tab:abl_mohe} shows that our proposed mixture ($\operatorname{DwConvFFN + FA-FFNs}$) achieves improvements in average MSE of $4.0\%$ on ETTh1 and $18.5\%$ on Traffic over the all-$\operatorname{MLP}$ version. 
    These results validate our hypothesis on the complementary expertise of the Mixture-of-Heterogeneous-Experts approach, specifically replacing traditional $\operatorname{MLP-FFN}$ experts with Fourier-based networks.

    \vspace{1pt}\textbf{Normalization Strategies.} We evaluate different normalization strategies, including homogeneous approaches with $\operatorname{LayerNorm}$~\cite{ba2016layer} or $\operatorname{RMSNorm}$ throughout the model (standard approach), and a mixed approach with a single-group group normalization (equivalent to a $\operatorname{LayerNorm}$ on channel dimensions) in the input patching and output projection, $\operatorname{RMSNorm}$ elsewhere (our proposal), inspired by CvT~\cite{wu2021cvt}. 
    \Tabref{tab:abl_norm} shows that the mixed approach achieves significantly lower MSE and MAE values than homogeneous normalizations on most benchmarks, with $9.9\%$ and $12.1\%$ improvements in MSE over all-$\operatorname{LayerNorm}$ and all-$\operatorname{RMSNorm}$, respectively, on ETTh1. However, the all-$\operatorname{RMSNorm}$ version presents a strong MSE on the Weather dataset.

    \vspace{1pt}\textbf{Projection Heads.} We compare the convolutional output decoder with traditional MLP-based heads. The convolutional design outperforms the MLP-based head by reasonable margins in most tests (\Tabref{tab:abl_norm}), validating that the convolutional locality's inductive bias is superior to dense linear projections for decoding. By respecting the patch boundary structure during upsampling, the convolutional head prevents overfitting on the output layer. 

    \vspace{1pt}\textbf{Exogenous Covariates.} Incorporating covariates through our multimodal cross-attention approach improves performance (\Tabref{tab:abl_norm}), particularly on datasets with lower dimensionality such as ETT.
    RoPE provides relative positions, but it lacks semantic context. A positional embedding informs the model that a time point $t$ is after $t-1$, but not that $t$ corresponds to a ``holiday,'' for example. Our cross-attention mechanism injects this semantic awareness directly, critical for modeling non-stationary shifts driven by human calendars.
    We observe a slight reduction in the advantage of injecting calendar information as data dimensionality increases, as in Traffic.
    Note that we only experimented with calendar information as covariates, which significantly improved robustness to non-stationarity in coarse datasets (e.g., ETTh1 and ETTh2). 
    However, we designed our cross-attention module as a general, extensible concept to handle information beyond just calendar marks.

\section{Conclusion}
\label{sec:conclusion}

    We presented \model, a novel transformer-based time series forecasting model that addresses the challenges of long-horizon multivariate prediction through tailored architectural innovations. 
    By integrating a shared time-domain expert and routed frequency-domain experts, our MoHE architecture enforces structural decomposition of global trends and local periodicities -- a combination that standard MLP-based experts fail to achieve. 
    Thus, \model establishes a multi-scale receptive field: global dependencies are resolved via attention, sequence-level continuity via the shared convolution, and local spectral details via Fourier experts.
    Additionally, the multimodal cross-attention mechanism integrates exogenous covariate embeddings, improving robustness to non-stationary dynamics. 
    \model achieved remarkable performance on multiple benchmarks, as validated through extensive ablation experiments. 
    Therefore, this work introduces advancements that position \model as a state-of-the-art solution for real-world time-series forecasting, providing a scalable and robust framework for diverse temporal applications.

\section*{Acknowledgments}
    
    This work was supported in part by the Paulo Pinheiro de Andrade Fellowship.
    The opinions, hypotheses, conclusions, or recommendations expressed in this material are the authors' responsibility and do not necessarily reflect the views of the funding agencies.

\section*{Impact Statement}

    This work advances long-term multivariate time-series forecasting, a capability that can improve decision-making in energy systems, climate modeling, supply chain management, and public health by increasing predictive accuracy and computational efficiency. 
    Although these improvements support positive social outcomes, such as reducing energy waste and improving disaster readiness, forecasting models also carry risks. In particular, they can reproduce or amplify biases present in historical data, which may be misapplied in automated high-stakes decision systems without adequate human oversight. 
    To mitigate these risks, we emphasize the need for rigorous validation, transparency in data provenance, and human oversight in deployment. 
    We are not aware of any malicious applications specific to this research beyond those common to forecasting systems, but we encourage practitioners to apply domain-appropriate safeguards before operational deployment.

\bibliography{ref}
\bibliographystyle{icml2026}

\newpage
\appendix
\onecolumn
\section{Experimental Details}
\label{apdx:details}

\subsection{Dataset Descriptions}
\label{apdx:datasets}

    We conduct long-term multivariate forecasting experiments on $7$ well-established real-world datasets to evaluate the performance of our \model, including: the ETT~\cite{zhou2021informer} series that contains four subsets with seven features related to power load of electricity transformers recorded during two years, where ETTh1 and ETTh2 are recorded hourly, and ETTm1 and ETTm2 are recorded every $15$ minutes; 
    Weather~\cite{wu2021autoformer} that includes $21$ meteorological features collected every $10$ minutes from the Max Planck Institute for Biogeochemistry; 
    ECL~\cite{wu2021autoformer} that contains hourly electricity consumption records from $321$ clients; and 
    Traffic~\cite{wu2021autoformer} records hourly road occupancy rates from $862$ sensors on San Francisco Bay freeways.
    These datasets are publicly available and have been extensively utilized for benchmarking time series forecasting models. The statistics of each dataset are provided in \Tabref{tab:dataset}.

    \begin{table}[htb]
    \centering
    \caption{Dataset descriptions. Dim denotes the number of variables. Dataset size refers to the number of time points and is organized into (Train, Validation, Test) splits.}
    \label{tab:dataset}
    \resizebox{0.81\columnwidth}{!}{
        \begin{small}
        \renewcommand{\multirowsetup}{\centering}
        \setlength{\tabcolsep}{3.8pt}
        \begin{tabular}{c|c|c|c|c|c|c}
        \toprule
            Task & Dataset & Dim & Dataset Size & Frequency & Forecastability & Information \\
        \midrule
        \multirow{6}{*}[-1.8em]{\begin{tabular}[c]{@{}c@{}}Long-term\\Forecasting\end{tabular}}
            & ETTh1 & 7 & (8545, 2881, 2881) & 1 Hour & 0.38 & Power Load \\
        \cmidrule{2-7}
            & ETTh2 & 7 & (8545, 2881, 2881) & 1 Hour & 0.45 & Power Load \\
        \cmidrule{2-7}
            & ETTm1 & 7 & (34465, 11521, 11521) & 15 Min & 0.46 & Power Load \\
        \cmidrule{2-7}
            & ETTm2 & 7 & (34465, 11521, 11521) & 15 Min & 0.55 & Power Load \\
        \cmidrule{2-7}
            & Weather & 21 & (36792, 5271, 10540) & 10 Min & 0.75  & Weather \\
        \cmidrule{2-7}
            & ECL & 321 & (18317, 2633, 5261) & 1 Hour & 0.77 & Electricity \\
        \cmidrule{2-7}
            & Traffic & 862 & (12185, 1757, 3509) & 1 Hour & 0.68  & Transportation \\
        \bottomrule
        \end{tabular}
        \end{small}
    }
    \end{table}

    Forecastability~\cite{goerg2013forecastable} is a measure of future uncertainty computed by one minus the entropy of the Fourier decomposition of time series. Higher values indicate better levels of predictability.

\subsection{Baseline Models}
\label{apdx:baselines}

    We select 15 advanced, well-known models as baselines for each experiment, representing the state-of-the-art in time series forecasting. These baselines include Transformer-based models such as TimeXer~\cite{wang2024timexer}, iTransformer~\cite{liu2023itransformer}, PatchTST~\cite{nie2022time}, Crossformer~\cite{zhang2022crossformer}, FEDformer~\cite{zhou2022fedformer}, Timer-XL~\cite{liu2024timerxl}, Time-MoE~\cite{shi2024time}, Moirai~\cite{woo2024moirai}, MOMENT~\cite{goswami2024moment}, and Chronos~\cite{ansari2024chronos}, 
    as well as SOFTS~\cite{han2024softs}, TimeMixer~\cite{wangtimemixer}, TiDE~\cite{das2023long}, and DLinear~\cite{zeng2023transformers}, which are based on MLP layers, and TimesNet~\cite{wu2023timesnet}, which is based on 2D convolutions.
    In particular, Timer-XL, TiDE, and TimeXer are recently published forecasters specifically designed to encode historical time series along with exogenous information. 

    SOFTS, TimeXer, iTransformer, TimeMixer, TimesNet, PatchTST, Crossformer, FEDformer, TiDE, and DLinear are in-domain (full-shot) time series models. 
    On the other hand, Timer-XL, Time-MoE, Moirai, MOMENT, and Chronos are large time-series foundation models pre-trained on multiple time-series datasets.
    In particular, Time-MoE is a mixture-of-experts Transformer with three model versions that scale to more than two billion parameters, pre-trained on a vast database comprising over $300$ billion time points~\cite{shi2024time}. 
    We report the official results from~\cite{liu2023itransformer,wang2024timexer,han2024softs,liu2024timer}.

\subsection{Metrics}
\label{apdx:metrics}

    We use the mean squared error (MSE) and mean absolute error (MAE) as evaluation metrics for the time-series forecasting experiments. A lower MSE or MAE indicates a better prediction. These metrics are calculated as follows:
    \begin{align*}
        \label{eq:metrics}
        \text{MSE} &= \frac{1}{H}\sum_{t=1}^H (\mathbf{x}_{t} - \widehat{\mathbf{x}}_{t})^2,
        &
        \text{MAE} &= \frac{1}{H}\sum_{t=1}^H|\mathbf{x}_{t} - \widehat{\mathbf{x}}_{t}|,
    \end{align*}
    where $\mathbf{x}_{t}, \widehat{\mathbf{x}}_{t} \in \mathbb{R}$ are the ground truth and the corresponding prediction of the $t$-th future time point. We further calculate the mean metric in the variable dimension for multivariate time series.

\subsection{Hyperparameter Settings}
\label{apdx:settings}

    With the settings defined in Sectio~\ref{subsec:settings}, we propose four model variants: \tmodel, with $0.3$ million activated parameters; \smodel, with $1.3$ million activated parameters; \basemodel, with $7.4$ million activated parameters; and \largemodel, with $21.4$ million activated parameters; all versions are designed for efficient inference on CPU architectures and significantly lighter compared to current state-of-the-art models in long-term time series forecasting~\cite{shi2024time,liu2024moirai,das2023decoder}. The detailed model configurations are in \Tabref{tab:model_size}.

    \begin{table}[htb]
        \caption{Summary of \model model configurations. Total Params can vary according to the data dimensionality.}
        \label{tab:model_size}%
        \centering
        \begin{small}
        \resizebox{0.94\columnwidth}{!}{
        \begin{tabular}{l@{\extracolsep{\fill}}ccccccccc}
            \toprule
                & Blocks & Q-Heads & KV-Heads & Experts & $K$ & \dmodel & \dff &
                Activated Params & Total Params \\
            \midrule
                \tmodel & 4 & 4 & 2 & 8 & 2 & 64 & 128 &
                0.3 $\mathrm{M}$ & 0.6 $\mathrm{M}$ \\ 
                \smodel & 4 & 4 & 2 & 8 & 2 & 128 & 256 &
                1.3 $\mathrm{M}$ & 2.5 $\mathrm{M}$ \\ 
                \basemodel & 6 & 8 & 4 & 8 & 2 & 256 & 512 &
                7.4 $\mathrm{M}$ & 14.5 $\mathrm{M}$ \\ 
                \largemodel & 8 & 12 & 6 & 8 & 2 & 384 & 768 &
                21.4 $\mathrm{M}$ & 42.7 $\mathrm{M}$ \\
            \bottomrule
        \end{tabular}%
        }
        \end{small}
    \end{table}%

    For all experiments, we define a standard base frequency of $10,000$ for Rotary Position Embeddings (RoPE). We apply DropPath~\cite{huang2016deep} with stochastic decay to the output of the attention and $\operatorname{MoHE}$ modules according to the depth of their Transformer blocks, with a maximum probability of $0.3$. We also apply Dropout~\cite{srivastava2014dropout} with probability $0.2$ to other encoder components. 
    Furthermore, we use $\operatorname{xavier\_uniform}$~\cite{glorot2010understanding} to initialize all the learnable weights of our model, except the Fourier Feed Forward networks in $\operatorname{MoHE}$ modules, which we initialize using normal distributions. 
    The reason for this choice lies in the trigonometric nature of the Fourier layers~\cite{dong2024fan}, which use explicit periodic $\cos$ and $\sin$ components to model periodic frequencies. 
    
    Widely used initializations such as $\operatorname{xavier\_}$ and $\operatorname{kaiming\_}$ are not ideal for periodic projections because they are designed to preserve variance for ReLU-style flows, but not for $\cos$ and $\sin$ operations. In practice, these initializations tend to produce distributions that are either too small or too structured, leading to periodic paths collapsing to near-constant behaviors.
    Therefore, initialization must place the weights of the Fourier layers in a moderate range, with $\mathcal{N}(0, 1)$ ensuring that during the first forward pass, the periodic features are well distributed to enable different phases (i.e., non-trivial), preventing them from collapsing near zero. 
    The code is publicly available on GitHub.

\subsection{Technical Details}
\label{apdx:technical}

    \textbf{Attention.} In \model, we implement grouped-query attention (GQA)~\cite{ainslie2023gqa} to optimize self-attention and cross-attention mechanisms, balancing computational efficiency and modeling capacity. Standard multi-head attention (MHA)~\cite{vaswani2017attention} allocates independent query (Q), key (K), and value (V) projections to each attention head, generating rich contextual information. The downside of MHA is the high memory bandwidth and computational costs, particularly for long sequences and autoregressive inference. 
    Multi-query attention (MQA)~\cite{shazeer2019fast} was proposed to mitigate attention costs by using a single K and V projection on multiple query heads, drastically reducing the memory footprint. However, this abrupt reduction in the K and V projections can degrade the model's capacity and training stability due to restricted representational expressivity, leading to a reduced performance compared to MHA~\cite{ainslie2023gqa}. 
    
\begin{figure}[htb]
        \centering
        \includegraphics[width=0.4\linewidth]{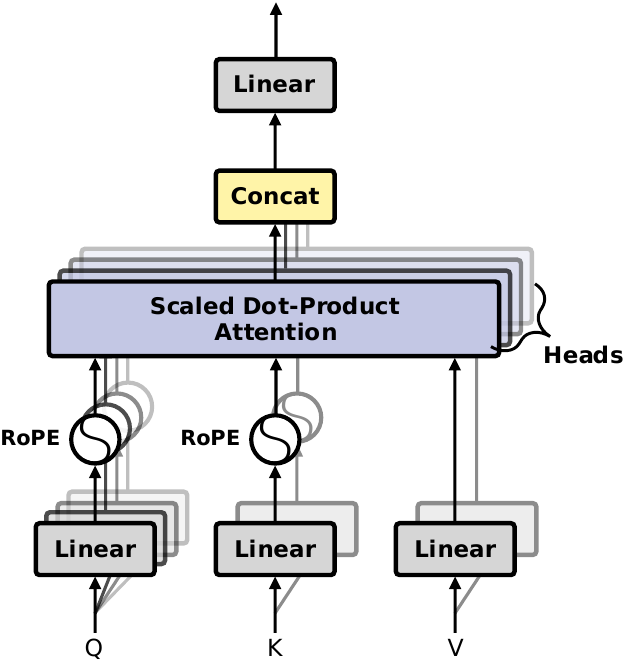}
        \caption{The grouped-query attention (GQA) mechanism with Rotary Position Embeddings (RoPE). Single key and value heads are shared for each group of query heads. In \model, we adopt a grouping factor of two query heads for each key/value head, i.e., Q-heads $= 2 \times$KV-heads.}
        \vspace{-4pt}
        \label{fig:gq_attn}
    \end{figure}

    GQA addresses the efficiency and capacity trade-off with a more general and flexible formulation of MHA that groups query heads into tunable clusters, each cluster sharing a single K and V projection, thereby reducing memory while preserving performance close to MHA~\cite{ainslie2023gqa}. 
    GQA has been successfully adopted in large-scale models, demonstrating improved efficiency with minimal loss of quality~\cite{grattafiori2024llama}. 
    For time series forecasting, long horizons require efficient attention. Thus, we adopt a light query grouping factor of 2 in \model, that is, Q-heads $= 2 \times$KV-heads (see Figure~\ref{fig:gq_attn} and \Tabref{tab:model_size}), which, combined with FlashAttention~\cite{dao2022flashattention}, reduces memory overhead while maintaining robust modeling capacity.

    \vspace{0.1cm}\textbf{MoHE Components.} Our Mixture-of-Heterogeneous-Experts approach consists of one shared expert designed to capture sequence-level temporal patterns and multiple routed experts designed to model patch-level periodic patterns. 
    All experts are randomly initialized. During training, the router learns to send similar patches to the same experts. Only selected experts will receive gradient updates and specialize in processing that type of patch.
    The routing mechanism operates independently in each $\operatorname{MoHE}$ layer along the model, creating a sophisticated network of specialization.
    For an input patch embedding $p$, the shared expert's contribution is gated by the weight $g_{N + 1,p}$, computed using the $\operatorname{Sigmoid}$ function (Equation~\ref{equ:mohe}). In contrast, the routed experts use the Topk $\operatorname{Softmax}$ routing weights $g_{i,p}$, as defined in Equations~\ref{equ:expert_score} and~\ref{equ:expert_gate}, retaining only the $K$ highest scores to ensure sparsity. 
    The final output combines the contributions of both shared and Topk experts to capture long-term patterns and local periodicity (see \Figref{fig:model_arch}c).
    The $\operatorname{MoHE}$ architecture activates only relevant parts of the model for each time patch, enabling model scaling without a corresponding increase in computation.
    It should be noted that we did not experiment with all convolutional experts because it does not make sense to process data units (single patches) using convolutions.

    \vspace{0.1cm}\textbf{Fourier-based Networks.} Multi-layer perceptron ($\operatorname{MLP}$) networks are widely used in machine learning and deep learning models, due to their general-purpose ability to approximate diverse functions. However, this general nature can hinder $\operatorname{MLPs}$ in accurately modeling patterns such as periodic signals~\cite{dong2024fan}. 
    In our $\operatorname{MoHE}$ modules, we replace the standard $\operatorname{MLP-FFNs}$ in routed experts with Fourier analysis networks ($\operatorname{FA-FFNs}$).  
    An $\operatorname{FA-FFN}$ is composed of two layers, following the typical inverted bottleneck design of Transformer blocks.
    Each layer is based on the Fourier series, which decomposes inputs into frequency-domain representations using sines and cosines, enhancing periodicity modeling. 
    The $\operatorname{FA-FFN}$ is defined as follows: 
    \vspace{5pt}
    \begin{equation}
        \text{FA-FFN}(\mathbf{x}) = \phi_{L2} \circ \phi_{L1} \circ \mathbf{x},
    \end{equation}
    with
    \begin{equation}
        \phi_l(\mathbf{x}) = [\cos(W^l_p\mathbf{x}) || \sin(W^l_p\mathbf{x}) || \sigma(W^l_{\bar{p}}\mathbf{x} + b^l_{\bar{p}})], \label{equ:fan}
    \end{equation}

    where $W_{p} \in \mathbb{R}^{d_{\text{model}} \times d_{\text{ff}}/4}, W_{\bar{p}} \in \mathbb{R}^{d_{\text{model}} \times d_{\text{ff}}/2}$, and $b_{\bar{p}} \in \mathbb{R}^{d_{\text{ff}}/2}$ are learnable parameters, $\sigma$ is a GELU activation function, and $||$ denotes concatenation.
    Note that an $\operatorname{MLP}$ layer is a special case of Equation~\ref{equ:fan} when the $W_p$ parameters are learned to be zero, which means that an $\operatorname{FA-FFN}$ is designed to model periodic signals, but can also retain general-purpose modeling capabilities as standard $\operatorname{FFN}$~\cite{dong2024fan}.

    \vspace{0.1cm}\textbf{Prediction Loss.} Time series forecasting models are often trained using the MSE loss.
    We deviate and use the Huber loss~\cite{huber1992robust,wen2019robusttrend}, which combines the advantages of the L1 and MSE losses to provide robustness to outliers, improving training stability~\cite{shi2024time}. For predicted time points $\hat{\mathbf{x}}_t$ and ground truth $\mathbf{x}_t$, the Huber loss is defined as:
    \vspace{5pt}
    \begin{equation}
        \mathcal{L}_{\text{pred}} \left( \mathbf{x}_t, \hat{\mathbf{x}}_t \right) = \begin{cases}
        0.5 \left( \mathbf{x}_t - \hat{\mathbf{x}}_t \right)^{2}, & \text{if } \left| \mathbf{x}_t - \hat{\mathbf{x}}_t \right| \leq \delta, \\
        \delta \times \left( \left| \mathbf{x}_t - \hat{\mathbf{x}}_t \right| - 0.5 \times \delta \right), & \text{otherwise},
        \end{cases}
    \end{equation}
    
    with $\delta$ as a hyperparameter that balances the scaled L1 and MSE losses. As demonstrated by~\cite{shi2024time}, time series models trained with Huber loss outperform those trained using only MSE loss due to the superior robustness of Huber loss in handling outliers. 

    \vspace{0.1cm}\textbf{Expert Balance Loss.} Sparse Mixture-of-Experts architectures, such as our MoHE, rely on automatically learned routing strategies that may suffer from load imbalance where a few experts dominate patch assignments. 
    As a consequence, the routing can collapse, leading to under-utilization of experts and reduced specialization~\cite{shi2024time,dai2024deepseekmoe}. 
    To avoid the risk of routing collapse, we incorporate the auxiliary expert balance loss proposed by~\cite{fedus2022switch}. This loss penalizes experts with high gating scores, promoting balanced loads among experts to prevent stronger experts from monopolizing patches during training, being computed as:
    \begin{equation}
        \mathcal{L}_{\text{aux}} = N \sum_{i=1}^{N}f_i r_i, 
        \label{equ:aux_loss}
    \end{equation}
    where $f_i$ denotes the fraction of input patches routed to expert $i$, and $r_i$ represents the average routing score allocated to expert $i$ by the gating mechanism. These quantities are formally defined as: 
    \begin{equation}
        f_i = \frac{1}{KP} \sum_{p=1}^{P} \mathbb{I} \left(\text{Time patch } p \text{ selects Expert } i \right), \quad r_i = \frac{1}{P} \sum_{p=1}^{P} s_{i,p},
    \end{equation}
    where $P$ is the number of input patches, $K$ is the number of experts selected per time patch, \(s_{i,p}\) is the expert routing probability (from $\operatorname{Softmax}$ score, see Equation~\ref{equ:expert_gate}), and $\mathbb{I}$ is the indicator function~\cite{dai2024deepseekmoe}.
    Therefore, the expert balance loss is calculated as the product of the fraction of patches, $f_i$, routed to each expert $i$, and the routing probability, $r_i$, thereby encouraging uniform expert utilization by assigning higher loss values to experts with higher routing probabilities.

    \vspace{0.1cm}\textbf{Output Head Architecture. } As illustrated in \Figref{fig:model_arch}d, we apply a final $\operatorname{RMSNorm}$ to the last Transformer block to improve stability and then forward the resulting patch embedding to the output decoder module, which is designed as follows. A single-layer $\operatorname{MLP}$ with dimensions \(d_{\text{model}} \times d_{\text{model}}\) receives the normalized embeddings, and a $\operatorname{ConvTranspose}$ layer unpatches each embedding to time points. The unpatched sequence is processed and projected by a convolutional block inspired by ConvNeXt's large kernel inverted bottleneck~\cite{convnet}, i.e., 
    (\textit{i}) a depthwise convolution with default kernel size of $7$ to focus on non-local temporal interactions at the sequence level, followed by a single-group $\operatorname{GroupNorm}$ along the channel dimension to normalize embeddings; 
    (\textit{ii}) a pointwise convolution reducing the dimension by a factor of $4$ followed by a GELU activation; and 
    (\textit{iii}) a final pointwise convolution projecting the embedding to a single dimension. Finally, channel-independent outputs are reshaped to the original data dimension, \(\mathbb{R}^{D \times H}\).

\begin{figure}[htb]
        \centering
        \includegraphics[width=0.82\linewidth]{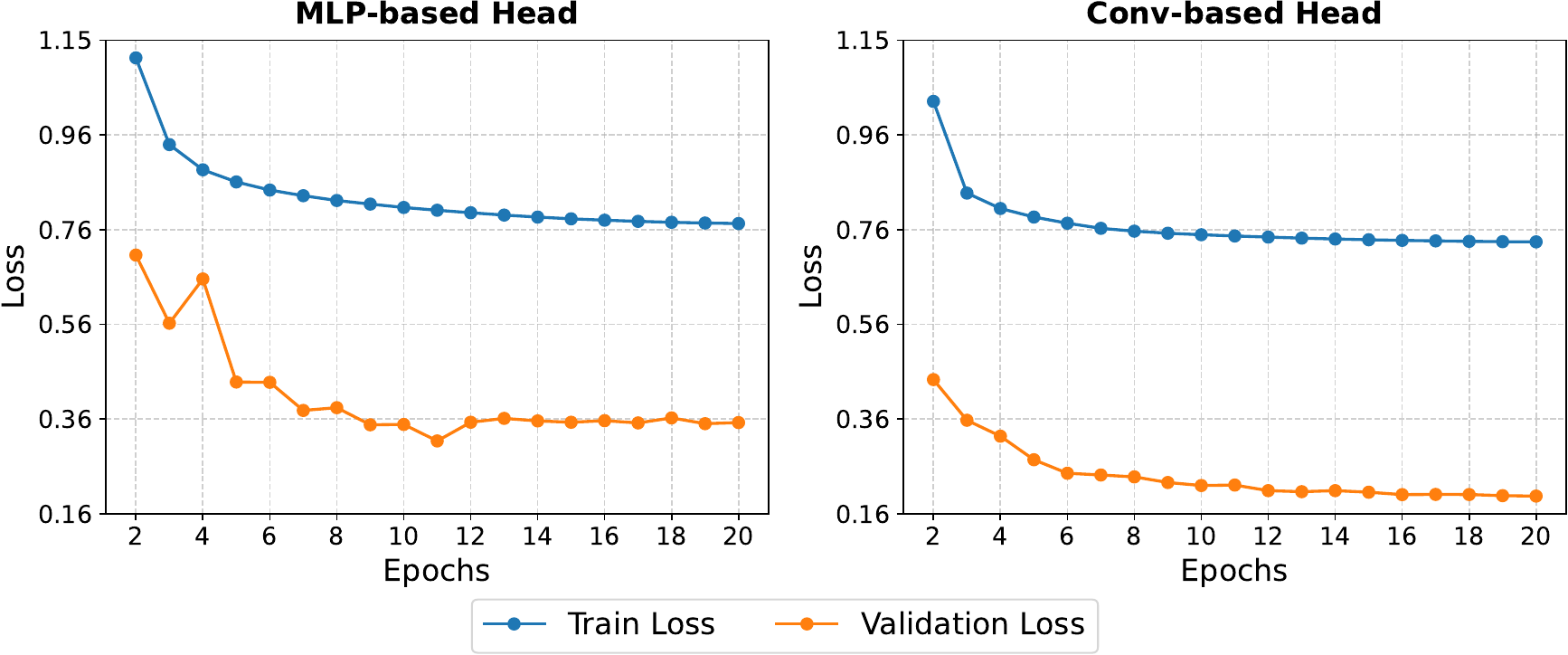}
        \caption{Training and validation loss curves of \model on Traffic data, comparing a conventional MLP-based projection head (left) with our convolutional head (right). We combine the Huber loss with the balanced loss for training (see Section~\ref{subsec:objective}) and use the MSE loss for validation.}
        \label{fig:heads}
    \end{figure}

    In \Figref{fig:heads}, we compare training and validation loss curves of \model on the Traffic dataset (\smodel, \(P = 12\), \(H_o = 24\), and 20 epochs of training), using a conventional MLP projection head (left) and our proposed convolutional decoder head (right). The convolutional head version achieves smoother training optimization compared to the oscillatory validation loss decrease observed in the MLP head version, leading \model to an average forecasting MSE of $0.406$, compared to $0.418$ for the MLP head (see \Tabref{tab:abl_norm}). Furthermore, the convolutional design is more parameter-efficient, reducing the total number of parameters by 42\% (from 4.7$\mathrm{M}$ parameters to 8.1$\mathrm{M}$ parameters for the MLP head version), indicating that the convolutional decoder improves generalization and reduces the model size.
    
\section{Full Experimental Results}
\label{apdx:results}

    In Table~\ref{tab:ltsf_full}, we provide the full results for each forecasting task in Table~\ref{tab:ltsf_avg_full}. We highlight the performance of \model in ultra-long-horizon multivariate forecasting tasks, that is, predicting $720$ future time points, where our model outperforms TimeXer by an $11.7\%$ reduction in MSE on the ETTh1 dataset and by an $11.2\%$ reduction in MSE on ETTm1, in addition to also outperforming TimeMixer by a $10.6\%$ reduction in MSE on the Weather data and TimeXer by a $3.8\%$ reduction in MSE on ECL. 
    These consistent gains underscore \model's robustness for extended forecasting.

    \begin{table}[!htb]
        \vspace{2pt}
        \centering
        \caption{Results of long-term multivariate forecasting experiments. Full-shot results are obtained from~\cite{liu2023itransformer,wang2024timexer,han2024softs}. {\bestres{Bold red}}: the best, \secondres{underlined blue}: the second best. $1^{\text{st}}$ Count represents the number of wins achieved by a model across all prediction lengths and datasets.}
        \label{tab:ltsf_full}
        \resizebox{\columnwidth}{!}{
        \renewcommand{\tabcolsep}{3pt}
        \begin{tabular}{cr|cc|cc|cc|cc|cc|cc|cc|cc|cc|cc|cc}
        \toprule
            \multicolumn{2}{c}{\scalebox{1.1}{\textbf{Models}}} & \multicolumn{2}{c}{\textbf{{\model}}} & \multicolumn{2}{c}{\textbf{SOFTS}} & \multicolumn{2}{c}{\textbf{TimeXer}} & \multicolumn{2}{c}{\textbf{iTransformer}} & \multicolumn{2}{c}{\textbf{TimeMixer}} &  \multicolumn{2}{c}{\textbf{TimesNet}} & \multicolumn{2}{c}{\textbf{PatchTST}} & \multicolumn{2}{c}{\textbf{Crossformer}} & \multicolumn{2}{c}{\textbf{TiDE}} & \multicolumn{2}{c}{\textbf{DLinear}} & \multicolumn{2}{c}{\textbf{FEDformer}} \\
    
            \cmidrule(lr){3-4} \cmidrule(lr){5-6}\cmidrule(lr){7-8} \cmidrule(lr){9-10}\cmidrule(lr){11-12}\cmidrule(lr){13-14}\cmidrule(lr){15-16}\cmidrule(lr){17-18}\cmidrule(lr){19-20} \cmidrule(lr){21-22} \cmidrule(lr){23-24}

            \multicolumn{2}{c}{\scalebox{1.1}{\textbf{Metrics $\downarrow$}}} & \textbf{MSE} & \textbf{MAE} & \textbf{MSE} & \textbf{MAE} & \textbf{MSE} & \textbf{MAE} & \textbf{MSE} & \textbf{MAE} & \textbf{MSE} & \textbf{MAE} & \textbf{MSE} & \textbf{MAE} & \textbf{MSE} & \textbf{MAE} & \textbf{MSE} & \textbf{MAE} & \textbf{MSE} & \textbf{MAE} & \textbf{MSE} & \textbf{MAE} & \textbf{MSE} & \textbf{MAE} \\
        \midrule
        \multirow{4}[1]{*}{ETTh1} 
            & 96 & \bestres{0.350} & \bestres{0.383} & 0.381 & \secondres{0.399} & 0.382 & 0.403 & 0.386 & 0.405 & \secondres{0.375} & 0.400 & 0.384 & 0.402 & 0.414 & 0.419 & 0.423 & 0.448 & 0.479 & 0.464 & 0.386 & 0.400 & 0.376 & 0.419 \\
            & 192 & \bestres{0.376} & \bestres{0.404} & 0.435 & 0.431 & 0.429 & 0.435 & 0.441 & 0.436 & 0.436 & \secondres{0.429} & 0.421 & \secondres{0.429} & 0.460 & 0.445 & 0.471 & 0.474 & 0.525 & 0.492 & 0.437 & 0.432 & \secondres{0.420} & 0.448 \\
            & 336 & \bestres{0.393} & \bestres{0.418} & 0.480 & 0.452 & 0.468 & \secondres{0.448} & 0.487 & 0.458 & 0.484 & 0.458 & 0.491 & 0.469 & 0.501 & 0.466 & 0.570 & 0.546 & 0.565 & 0.515 & 0.481 & 0.459 & \secondres{0.459} & 0.465 \\
            & 720 & \bestres{0.414} & \bestres{0.441} & 0.499 & 0.488 & \secondres{0.469} & \secondres{0.461} & 0.503 & 0.491 & 0.498 & 0.482 & 0.521 & 0.500 & 0.500 & 0.488 & 0.653 & 0.621 & 0.594 & 0.558 & 0.519 & 0.516 & 0.506 & 0.507 \\
            \rowcolor{tabhighlight}
            & {\textbf{Avg.}} & \bestres{0.383} & \bestres{0.412} & 0.449 & 0.442 & \secondres{0.437} & \secondres{0.437} & 0.454 & 0.447 & 0.448 & 0.442 & 0.454 & 0.450 & 0.469 & 0.454 & 0.529 & 0.522 & 0.540 & 0.507 & 0.455 & 0.451 & 0.440 & 0.459 \\
        \midrule
        \multirow{4}[0]{*}{ETTh2} 
            & 96 & \bestres{0.278} & \bestres{0.332} & 0.297 & 0.347 & \secondres{0.286} & \secondres{0.338} & 0.297 & 0.349 & 0.289 & 0.341 & 0.340 & 0.374 & 0.302 & 0.348 & 0.745 & 0.584 & 0.400 & 0.440 & 0.333 & 0.387 & 0.358 & 0.397 \\
            & 192 & \bestres{0.345} & \bestres{0.374} & 0.373 & 0.394 & \secondres{0.363} & \secondres{0.389} & 0.380 & 0.400 & 0.372 & 0.392 & 0.402 & 0.414 & 0.388 & 0.400 & 0.877 & 0.656 & 0.528 & 0.509 & 0.477 & 0.476 & 0.429 & 0.439 \\
            & 336 & \bestres{0.376} & \bestres{0.400} & 0.410 & 0.426 & 0.414 & 0.423 & 0.428 & 0.432 & \secondres{0.386} & \secondres{0.414} & 0.452 & 0.541 & 0.426 & 0.433 & 1.043 & 0.731 & 0.643 & 0.571 & 0.594 & 0.541 & 0.496 & 0.487 \\
            & 720 & \bestres{0.392} & \bestres{0.421} & 0.411 & 0.433 & \secondres{0.408} & \secondres{0.432} & 0.427 & 0.445 & 0.412 & 0.434 & 0.462 & 0.657 & 0.431 & 0.446 & 1.104 & 0.763 & 0.874 & 0.679 & 0.831 & 0.657 & 0.463 & 0.474 \\
            \rowcolor{tabhighlight}
            & {\textbf{Avg.}} & \bestres{0.348} & \bestres{0.382} & 0.373 & 0.400 & 0.367 & 0.396 & 0.383 & 0.406 & \secondres{0.364} & \secondres{0.395} & 0.414 & 0.496 & 0.387 & 0.407 & 0.942 & 0.683 & 0.611 & 0.549 & 0.558 & 0.515 & 0.436 & 0.449 \\
        \midrule
        \multirow{4}[0]{*}{ETTm1} 
            & 96 & \bestres{0.276} & \bestres{0.327} & 0.325 & 0.361 & \secondres{0.318} & \secondres{0.356} & 0.334 & 0.368 & 0.320 & 0.357 & 0.338 & 0.375 & 0.329 & 0.367 & 0.404 & 0.426 & 0.364 & 0.387 & 0.345 & 0.372 & 0.379 & 0.419 \\
            & 192 & \bestres{0.313} & \bestres{0.354} & 0.375 & 0.389 & 0.362 & 0.383 & 0.377 & 0.391 & \secondres{0.361} & \secondres{0.381} & 0.374 & 0.387 & 0.367 & 0.385 & 0.450 & 0.451 & 0.398 & 0.404 & 0.380 & 0.389 & 0.426 & 0.441 \\
            & 336 & \bestres{0.343} & \bestres{0.376} & 0.405 & 0.412 & 0.395 & 0.407 & 0.426 & 0.420 & \secondres{0.390} & \secondres{0.404} & 0.410 & 0.411 & 0.399 & 0.410 & 0.532 & 0.515 & 0.428 & 0.425 & 0.413 & 0.413 & 0.445 & 0.459 \\
            & 720 & \bestres{0.401} & \bestres{0.410} & 0.466 & 0.447 & \secondres{0.452} & 0.441 & 0.491 & 0.459 & 0.454 & 0.441 & 0.478 & 0.450 & 0.454 & \secondres{0.439} & 0.666 & 0.589 & 0.487 & 0.461 & 0.474 & 0.453 & 0.543 & 0.490 \\
            \rowcolor{tabhighlight}
            & {\textbf{Avg.}} & \bestres{0.333} & \bestres{0.367} & 0.393 & 0.403 & 0.382 & 0.397 & 0.407 & 0.410 & \secondres{0.381} & \secondres{0.395} & 0.400 & 0.405 & 0.387 & 0.400 & 0.513 & 0.495 & 0.419 & 0.419 & 0.403 & 0.406 & 0.448 & 0.452 \\
        \midrule
        \multirow{4}[0]{*}{ETTm2} 
            & 96 & \bestres{0.164} & \bestres{0.249} & 0.180 & 0.261 & \secondres{0.171} & \secondres{0.256} & 0.180 & 0.264 & 0.175 & 0.258 & 0.187 & 0.267 & 0.175 & 0.259 & 0.287 & 0.366 & 0.207 & 0.305 & 0.193 & 0.292 & 0.203 & 0.287 \\
            & 192 & \bestres{0.222} & \bestres{0.288} & 0.246 & 0.306 & \secondres{0.237} & \secondres{0.299} & 0.250 & 0.309 & \secondres{0.237} & \secondres{0.299} & 0.249 & 0.309 & 0.241 & 0.302 & 0.414 & 0.492 & 0.290 & 0.364 & 0.284 & 0.362 & 0.269 & 0.328 \\
            & 336 & \bestres{0.275} & \bestres{0.323} & 0.319 & 0.352 & \secondres{0.296} & \secondres{0.338} & 0.311 & 0.348 & 0.298 & 0.340 & 0.321 & 0.351 & 0.305 & 0.343 & 0.597 & 0.542 & 0.377 & 0.422 & 0.369 & 0.427 & 0.325 & 0.366 \\
            & 720 & \bestres{0.361} & \bestres{0.378} & 0.405 & 0.401 & 0.392 & \secondres{0.394} & 0.412 & 0.407 & \secondres{0.391} & 0.396 & 0.408 & 0.403 & 0.402 & 0.400 & 1.730 & 1.042 & 0.558 & 0.524 & 0.554 & 0.522 & 0.421 & 0.415 \\
            \rowcolor{tabhighlight}
            & {\textbf{Avg.}} & \bestres{0.256} & \bestres{0.310} & 0.287 & 0.330 & \secondres{0.274} & \secondres{0.322} & 0.288 & 0.332 & 0.275 & 0.323 & 0.291 & 0.332 & 0.281 & 0.326 & 0.757 & 0.610 & 0.358 & 0.403 & 0.350 & 0.400 & 0.304 & 0.349 \\
        \midrule
        \multirow{4}[0]{*}{Weather} 
            & 96 & \bestres{0.141} & \bestres{0.184} & 0.166 & 0.208 & \secondres{0.157} & \secondres{0.205} & 0.174 & 0.214 & 0.163 & 0.209 & 0.172 & 0.220 & 0.177 & 0.218 & 0.158 & 0.230 & 0.202 & 0.261 & 0.196 & 0.255 & 0.217 & 0.296 \\
            & 192 & \bestres{0.184} & \bestres{0.227} & 0.217 & 0.253 & \secondres{0.204} & \secondres{0.247} & 0.221 & 0.254 & 0.208 & 0.250 & 0.219 & 0.261 & 0.225 & 0.259 & 0.206 & 0.277 & 0.242 & 0.298 & 0.237 & 0.296 & 0.276 & 0.336 \\
            & 336 & \bestres{0.235} & \bestres{0.268} & 0.282 & 0.300 & 0.261 & 0.290 & 0.278 & 0.296 & \secondres{0.251} & \secondres{0.287} & 0.280 & 0.306 & 0.278 & 0.297 & 0.272 & 0.335 & 0.287 & 0.335 & 0.283 & 0.335 & 0.339 & 0.380 \\
            & 720 & \bestres{0.303} & \bestres{0.318} & 0.356 & 0.351 & 0.340 & \secondres{0.341} & 0.358 & 0.349 & \secondres{0.339} & \secondres{0.341} & 0.365 & 0.359 & 0.354 & 0.348 & 0.398 & 0.418 & 0.351 & 0.386 & 0.345 & 0.381 & 0.403 & 0.428 \\
            \rowcolor{tabhighlight}
            & {\textbf{Avg.}} & \bestres{0.216} & \bestres{0.249} & 0.255 & 0.278 & 0.241 & \secondres{0.271} & 0.258 & 0.278 & \secondres{0.240} & \secondres{0.271} & 0.259 & 0.286 & 0.259 & 0.281 & 0.258 & 0.315 & 0.270 & 0.320 & 0.265 & 0.316 & 0.308 & 0.360 \\
        \midrule
        \multirow{4}[0]{*}{ECL}
            & 96 & \bestres{0.125} & \bestres{0.218} & 0.143 & \secondres{0.233} & \secondres{0.140} & 0.242 & 0.148 & 0.240 & 0.153 & 0.247 & 0.168 & 0.272 & 0.181 & 0.270 & 0.219 & 0.314 & 0.237 & 0.329 & 0.197 & 0.282 & 0.193 & 0.308 \\
            & 192 & \bestres{0.144} & \bestres{0.236} & 0.158 & \secondres{0.248} & \secondres{0.157} & 0.256 & 0.162 & 0.253 & 0.166 & 0.256 & 0.184 & 0.289 & 0.188 & 0.274 & 0.231 & 0.322 & 0.236 & 0.330 & 0.196 & 0.285 & 0.201 & 0.315 \\
            & 336 & \bestres{0.161} & \bestres{0.256} & 0.178 & \secondres{0.269} & \secondres{0.176} & 0.275 & 0.178 & \secondres{0.269} & 0.185 & 0.277 & 0.198 & 0.300 & 0.204 & 0.293 & 0.246 & 0.337 & 0.249 & 0.344 & 0.209 & 0.301 & 0.214 & 0.329 \\
            & 720 & \bestres{0.203} & \bestres{0.295} & 0.218 & \secondres{0.305} & \secondres{0.211} & 0.306 & 0.225 & 0.317 & 0.225 & 0.310 & 0.220 & 0.320 & 0.246 & 0.324 & 0.280 & 0.363 & 0.284 & 0.373 & 0.245 & 0.333 & 0.246 & 0.355 \\
            \rowcolor{tabhighlight}
            & {\textbf{Avg.}} & \bestres{0.158} & \bestres{0.251} & 0.174 & \secondres{0.264} & \secondres{0.171} & 0.270 & 0.178 & 0.270 & 0.182 & 0.272 & 0.192 & 0.295 & 0.205 & 0.290 & 0.244 & 0.334 & 0.251 & 0.344 & 0.212 & 0.300 & 0.214 & 0.327 \\
        \midrule
        \multirow{4}[0]{*}{Traffic} 
            & 96 & \bestres{0.351} & \bestres{0.236} & \secondres{0.376} & \secondres{0.251} & 0.428 & 0.271 & 0.395 & 0.268 & 0.462 & 0.285 & 0.593 & 0.321 & 0.462 & 0.295 & 0.522 & 0.290 & 0.805 & 0.493 & 0.650 & 0.396 & 0.587 & 0.366 \\
            & 192 & \bestres{0.372} & \bestres{0.247} & \secondres{0.398} & \secondres{0.261} & 0.448 & 0.282 & 0.417 & 0.276 & 0.473 & 0.296 & 0.617 & 0.336 & 0.466 & 0.296 & 0.530 & 0.293 & 0.756 & 0.474 & 0.598 & 0.370 & 0.604 & 0.373 \\
            & 336 & \bestres{0.391} & \bestres{0.257} & \secondres{0.415} & \secondres{0.269} & 0.473 & 0.289 & 0.433 & 0.283 & 0.498 & 0.296 & 0.629 & 0.336 & 0.482 & 0.304 & 0.558 & 0.305 & 0.762 & 0.477 & 0.605 & 0.373 & 0.621 & 0.383 \\
            & 720 & \bestres{0.440} & \bestres{0.284} & \secondres{0.447} & \secondres{0.287} & 0.516 & 0.307 & 0.467 & 0.302 & 0.506 & 0.313 & 0.640 & 0.350 & 0.514 & 0.322 & 0.589 & 0.328 & 0.719 & 0.449 & 0.645 & 0.394 & 0.626 & 0.382 \\
            \rowcolor{tabhighlight}
            & {\textbf{Avg.}} & \bestres{0.388} & \bestres{0.256} & \secondres{0.409} & \secondres{0.267} & 0.466 & 0.287 & 0.428 & 0.282 & 0.484 & 0.297 & 0.620 & 0.336 & 0.481 & 0.304 & 0.550 & 0.304 & 0.760 & 0.473 & 0.625 & 0.383 & 0.609 & 0.376 \\
        \midrule
        \multicolumn{2}{c|}{\scalebox{1.1}{\textbf{Average}}}
            & \bestres{0.297} & \bestres{0.318} & \secondres{0.334} & 0.341 & \secondres{0.334} & \secondres{0.340} & 0.342 & 0.346 & 0.339 & 0.342 & 0.376 & 0.371 & 0.353 & 0.352 & 0.542 & 0.466 & 0.458 & 0.431 & 0.410 & 0.396 & 0.394 & 0.396 \\
        \midrule
        \rowcolor{tabhighlight}\multicolumn{2}{c}{\textbf{{$1^{\text{st}}$ Count}}} 
            & \multicolumn{2}{c}{72} & \multicolumn{2}{c}{0} & \multicolumn{2}{c}{0} & \multicolumn{2}{c}{0} & \multicolumn{2}{c}{0} & \multicolumn{2}{c}{0} & \multicolumn{2}{c}{0} & \multicolumn{2}{c}{0} &  \multicolumn{2}{c}{0} & \multicolumn{2}{c}{0} & \multicolumn{2}{c}{0} \\
        \bottomrule
        \end{tabular}%
        }
    \end{table}%

    In Table~\ref{tab:ltsf_additional}, we present additional comparisons with time series foundation models to provide a more comprehensive evaluation. 
    Specifically, we include Timer-XL, and all versions of Time-MoE, Moirai, MOMENT, and Chronos (see Section~\ref{apdx:baselines}). 
    It should be noted that Time-MoE and Moirai achieve strong results on the ETTh1 and ETTh2 datasets. 
    These datasets are small and coarse-grained, making them prone to overfitting even with models of moderate size~\cite{nie2022time}, suggesting that pre-training on large-scale time series is a valuable approach to enhance performance in such contexts.

    In particular, Time-MoE, a foundation model pre-trained on a vast time-series dataset, outperforms \model across a few horizons. 
    However, \model outperforms Time-MoE in most benchmarks, including a $15.6\%$ reduction in average MSE on the Weather data relative to Time-MoE\textsubscript{ultra}. 
    We highlight that \model is considerably lighter, demonstrating that combining carefully designed architectures to capture heterogeneous patterns from time series enables \model to achieve state-of-the-art performance efficiently.
    The general results in Table~\ref{tab:ltsf_additional} indicate that our model performs consistently against such established approaches, including the $2.4$-billion-parameter Time-MoE\textsubscript{ultra}.

    \begin{table}[!htb]
        \centering
        \caption{Additional results of long-term multivariate forecasting against large foundation model baselines. Results are obtained from~\cite{liu2024timer}. {\bestres{Bold red}}: the best, \secondres{underlined blue}: the second best. $1^{\text{st}}$ Count represents the number of wins achieved by a model across all prediction lengths and datasets.}
        \label{tab:ltsf_additional}
        \resizebox{\columnwidth}{!}{
        \begin{threeparttable}
        \renewcommand{\tabcolsep}{3pt}
        \begin{tabular}{cr|cc|cc|cc|cc|cc|cc|cc|cc|cc|cc|cc}
        \toprule
            \multicolumn{2}{c}{\scalebox{1.1}{\textbf{Models}}} & 
            \multicolumn{2}{c}{\textbf{{\model}}} & 
            \multicolumn{2}{c}{\textbf{Timer-XL}\textsubscript{Base}} &
            \multicolumn{2}{c}{\textbf{Time-MoE}\textsubscript{Base}} &
            \multicolumn{2}{c}{\textbf{Time-MoE}\textsubscript{Large}} &
            \multicolumn{2}{c}{\textbf{Time-MoE}\textsubscript{Ultra}} &
            \multicolumn{2}{c}{\textbf{Moirai}\textsubscript{Small}} &
            \multicolumn{2}{c}{\textbf{Moirai}\textsubscript{Base}} &
            \multicolumn{2}{c}{\textbf{Moirai}\textsubscript{Large}} &
            \multicolumn{2}{c}{\textbf{MOMENT}} &
            \multicolumn{2}{c}{\textbf{Chronos}\textsubscript{Base}} &
            \multicolumn{2}{c}{\textbf{Chronos}\textsubscript{Large}} \\
    
            \cmidrule(lr){3-4} \cmidrule(lr){5-6}\cmidrule(lr){7-8} \cmidrule(lr){9-10}\cmidrule(lr){11-12}\cmidrule(lr){13-14}\cmidrule(lr){15-16}\cmidrule(lr){17-18}\cmidrule(lr){19-20} \cmidrule(lr){21-22} \cmidrule(lr){23-24}

            \multicolumn{2}{c}{\scalebox{1.1}{\textbf{Metrics $\downarrow$}}} & \textbf{MSE} & \textbf{MAE} & \textbf{MSE} & \textbf{MAE} & \textbf{MSE} & \textbf{MAE} & \textbf{MSE} & \textbf{MAE} & \textbf{MSE} & \textbf{MAE} & \textbf{MSE} & \textbf{MAE} & \textbf{MSE} & \textbf{MAE} & \textbf{MSE} & \textbf{MAE} & \textbf{MSE} & \textbf{MAE} & \textbf{MSE} & \textbf{MAE} & \textbf{MSE} & \textbf{MAE} \\
        \midrule
        \multirow{4}[1]{*}{ETTh1}
            & 96 & \secondres{0.350} & 0.383 & 0.369 & 0.391 & 0.357 & \secondres{0.381} & \secondres{0.350} & 0.382 & \bestres{0.349} & \bestres{0.379} & 0.401 & 0.402 & 0.376 & 0.392 & 0.381 & 0.388 & 0.688 & 0.557 & 0.440 & 0.393 & 0.441 & 0.390 \\
            & 192 & \bestres{0.376} & \bestres{0.404} & 0.405 & 0.413 & \secondres{0.384} & \bestres{0.404} & 0.388 & \secondres{0.412} & 0.395 & 0.413 & 0.435 & 0.421 & 0.412 & 0.413 & 0.434 & 0.415 & 0.688 & 0.560 & 0.492 & 0.426 & 0.502 & 0.524 \\
            & 336 & \bestres{0.393} & \bestres{0.418} & 0.418 & \secondres{0.423} & \secondres{0.411} & 0.434 & \secondres{0.411} & 0.430 & 0.447 & 0.453 & 0.438 & 0.434 & 0.433 & 0.428 & 0.485 & 0.445 & 0.675 & 0.563 & 0.550 & 0.462 & 0.576 & 0.467 \\
            & 720 & \bestres{0.414} & \bestres{0.441} & \secondres{0.423} & \bestres{0.441} & 0.449 & 0.477 & 0.427 & 0.455 & 0.457 & 0.462 & 0.439 & 0.454 & 0.447 & \secondres{0.444} & 0.611 & 0.510 & 0.683 & 0.585 & 0.882 & 0.591 & 0.835 & 0.583 \\
            \rowcolor{tabhighlight}
            & {\textbf{Avg.}} & \bestres{0.383} & \bestres{0.412} & 0.404 & \secondres{0.417} & 0.400 & 0.424 & \secondres{0.394} & 0.419 & 0.412 & 0.426 & 0.428 & 0.427 & 0.417 & 0.419 & 0.480 & 0.439 & 0.683 & 0.566 & 0.591 & 0.468 & 0.588 & 0.466 \\
        \midrule
        \multirow{4}[0]{*}{ETTh2}
            & 96 & \bestres{0.278} & \secondres{0.332} & \secondres{0.283} & 0.342 & 0.305 & 0.359 & 0.302 & 0.354 & 0.292 & 0.352 & 0.297 & 0.336 & 0.294 & \bestres{0.330} & 0.296 & \bestres{0.330} & 0.342 & 0.396 & 0.308 & 0.343 & 0.320 & 0.345 \\
            & 192 & \secondres{0.345} & \secondres{0.374} & \bestres{0.340} & 0.379 & 0.351 & 0.386 & 0.364 & 0.385 & 0.347 & 0.379 & 0.368 & 0.381 & 0.365 & 0.375 & 0.361 & \bestres{0.371} & 0.354 & 0.402 & 0.384 & 0.392 & 0.406 & 0.399 \\
            & 336 & 0.376 & \secondres{0.400} &
            \secondres{0.366} & \secondres{0.400} & 0.391 & 0.418 & 0.417 & 0.425 & 0.406 & 0.419 & 0.370 & 0.393 & 0.376 & \bestres{0.390} & 0.390 & \bestres{0.390} & \bestres{0.356} & 0.407 & 0.429 & 0.430 & 0.492 & 0.453 \\
            & 720 & \bestres{0.392} & \secondres{0.421} & 0.397 & 0.431 & 0.419 & 0.454 & 0.537 & 0.496 & 0.439 & 0.447 & 0.411 & 0.426 & 0.416 & 0.433 & 0.423 & \bestres{0.418} & \secondres{0.395} & 0.434 & 0.501 & 0.477 & 0.603 & 0.511 \\
            \rowcolor{tabhighlight}
            & {\textbf{Avg.}} & \secondres{0.348} & \secondres{0.382} &
            \bestres{0.347} & 0.388 & 0.366 & 0.404 & 0.405 & 0.415 & 0.371 & 0.399 & 0.361 & 0.384 & 0.362 & \secondres{0.382} & 0.367 & \bestres{0.377} & 0.361 & 0.409 & 0.405 & 0.410 & 0.455 & 0.427 \\
        \midrule
        \multirow{4}[0]{*}{ETTm1}
            & 96 & \bestres{0.276} & \bestres{0.327} & 0.317 & 0.356 & 0.338 & 0.368 & 0.309 & 0.357 & \secondres{0.281} & \secondres{0.341} & 0.418 & 0.392 & 0.363 & 0.356 & 0.380 & 0.361 & 0.654 & 0.527 & 0.454 & 0.408 & 0.457 & 0.403 \\
            & 192 & \secondres{0.313} & \bestres{0.354} & 0.358 & 0.381 & 0.353 & 0.388 & 0.346 & 0.381 & \bestres{0.305} & \secondres{0.358} & 0.431 & 0.405 & 0.388 & 0.375 & 0.412 & 0.383 & 0.662 & 0.532 & 0.567 & 0.477 & 0.530 & 0.450 \\
            & 336 & \bestres{0.343} & \bestres{0.376} & 0.386 & 0.401 & 0.381 & 0.413 & 0.373 & 0.408 & \secondres{0.369} & 0.395 & 0.433 & 0.412 & 0.416 & \secondres{0.392} & 0.436 & 0.400 & 0.672 & 0.537 & 0.662 & 0.525 & 0.577 & 0.481 \\
            & 720 & \bestres{0.401} & \bestres{0.410} & \secondres{0.430} & 0.431 & 0.504 & 0.493 & 0.475 & 0.477 & 0.469 & 0.472 & 0.462 & 0.432 & 0.460 & \secondres{0.418} & 0.462 & 0.420 & 0.692 & 0.551 & 0.900 & 0.591 & 0.660 & 0.526 \\
            \rowcolor{tabhighlight}
            & {\textbf{Avg.}} & \bestres{0.333} & \bestres{0.367} & 0.373 & 0.392 & 0.394 & 0.415 & 0.376 & 0.405 & \secondres{0.356} & 0.391 & 0.436 & 0.410 & 0.406 & \secondres{0.385} & 0.422 & 0.391 & 0.670 & 0.536 & 0.645 & 0.500 & 0.555 & 0.465 \\
        \midrule
        \multirow{4}[0]{*}{ETTm2}
            & 96 & \bestres{0.164} & \bestres{0.249} & \secondres{0.189} & 0.277 & 0.201 & 0.291 & 0.197 & 0.286 & 0.198 & 0.288 & 0.214 & 0.288 & 0.205 & 0.273 & 0.211 & 0.274 & 0.260 & 0.335 & 0.199 & 0.274 & 0.197 & \secondres{0.271} \\
            & 192 & \bestres{0.222} & \bestres{0.288} & 0.241 & 0.315 & 0.258 & 0.334 & 0.250 & 0.322 & \secondres{0.235} & \secondres{0.312} & 0.284 & 0.332 & 0.275 & 0.316 & 0.281 & 0.318 & 0.289 & 0.350 & 0.261 & 0.322 & 0.254 & 0.314 \\
            & 336 & \bestres{0.275} & \bestres{0.323} & \secondres{0.286} & \secondres{0.348} & 0.324 & 0.373 & 0.337 & 0.375 & 0.293 & \secondres{0.348} & 0.331 & 0.362 & 0.329 & 0.350 & 0.341 & 0.355 & 0.324 & 0.369 & 0.326 & 0.366 & 0.313 & 0.353 \\
            & 720 & \bestres{0.361} & \bestres{0.378} & \secondres{0.375} & \secondres{0.402} & 0.488 & 0.464 & 0.480 & 0.461 & 0.427 & 0.428 & 0.402 & 0.408 & 0.437 & 0.411 & 0.485 & 0.428 & 0.394 & 0.409 & 0.455 & 0.439 & 0.416 & 0.415 \\
            \rowcolor{tabhighlight}
            & {\textbf{Avg.}} & \bestres{0.256} & \bestres{0.310} & \secondres{0.273} & \secondres{0.336} & 0.317 & 0.365 & 0.316 & 0.361 & 0.288 & 0.344 & 0.307 & 0.347 & 0.311 & 0.337 & 0.329 & 0.343 & 0.316 & 0.365 & 0.310 & 0.350 & 0.295 & 0.338 \\
        \midrule
        \multirow{4}[0]{*}{Weather}
            & 96 & \bestres{0.141} & \bestres{0.184} & 0.171 & 0.225 & 0.160 & 0.214 & 0.159 & 0.213 & \secondres{0.157} & \secondres{0.211} & 0.198 & 0.222 & 0.220 & 0.217 & 0.199 & \secondres{0.211} & 0.243 & 0.255 & 0.203 & 0.238 & 0.194 & 0.235 \\
            & 192 & \bestres{0.184} & \bestres{0.227} & 0.221 & 0.271 & 0.210 & 0.260 & 0.215 & 0.266 & \secondres{0.208} & 0.256 & 0.247 & 0.265 & 0.271 & 0.259 & 0.246 & \secondres{0.251} & 0.278 & 0.329 & 0.256 & 0.290 & 0.249 & 0.285 \\
            & 336 & \bestres{0.235} & \bestres{0.268} & 0.274 & 0.311 & 0.274 & 0.309 & 0.291 & 0.322 & \secondres{0.255} & \secondres{0.290} & 0.283 & 0.303 & 0.286 & 0.297 & 0.274 & 0.291 & 0.306 & 0.346 & 0.314 & 0.336 & 0.302 & 0.327 \\
            & 720 & \bestres{0.303} & \bestres{0.318} & 0.356 & 0.370 & 0.418 & 0.405 & 0.415 & 0.400 & 0.405 & 0.397 & 0.373 & 0.354 & 0.373 & 0.354 & \secondres{0.337} & \secondres{0.340} & 0.350 & 0.374 & 0.397 & 0.396 & 0.372 & 0.378 \\
            \rowcolor{tabhighlight}
            & {\textbf{Avg.}} & \bestres{0.216} & \bestres{0.249} & \secondres{0.256} & 0.294 & 0.265 & 0.297 & 0.270 & 0.300 & \secondres{0.256} & 0.288 & 0.275 & 0.286 & 0.287 & 0.281 & 0.264 & \secondres{0.273} & 0.294 & 0.326 & 0.292 & 0.315 & 0.279 & 0.306 \\
        \midrule
        \multirow{4}[0]{*}{ECL}
            & 96 & \bestres{0.125} & \bestres{0.218} & \secondres{0.141} & 0.237 & -- & -- & -- & -- & -- & -- & 0.189 & 0.280 & 0.160 & 0.250 & 0.153 & 0.241 & 0.745 & 0.680 & 0.154 & 0.231 & 0.152 & \secondres{0.229} \\
            & 192 & \bestres{0.144} & \bestres{0.236} & \secondres{0.159} & 0.254 & -- & -- & -- & -- & -- & -- & 0.205 & 0.292 & 0.175 & 0.263 & 0.169 & 0.255 & 0.755 & 0.683 & 0.179 & 0.254 & 0.172 & \secondres{0.250} \\
            & 336 & \bestres{0.161} & \bestres{0.256} & \secondres{0.177} & \secondres{0.272} & -- & -- & -- & -- & -- & -- & 0.221 & 0.307 & 0.187 & 0.277 & 0.187 & 0.273 & 0.766 & 0.687 & 0.214 & 0.284 & 0.203 & 0.276 \\
            & 720 & \bestres{0.203} & \bestres{0.295} & \secondres{0.219} & \secondres{0.308} & -- & -- & -- & -- & -- & -- & 0.258 & 0.335 & 0.228 & 0.309 & 0.237 & 0.313 & 0.794 & 0.696 & 0.311 & 0.346 & 0.289 & 0.337 \\
            \rowcolor{tabhighlight}
            & {\textbf{Avg.}} & \bestres{0.158} & \bestres{0.251} & \secondres{0.174} & 0.278 & -- & -- & -- & -- & -- & -- & 0.218 & 0.303 & 0.187 & 0.274 & 0.186 & \secondres{0.270} & 0.765 & 0.686 & 0.214 & 0.278 & 0.204 & 0.273 \\
        \midrule
        \multicolumn{2}{c|}{\scalebox{1.1}{\textbf{Average}}}
            & \bestres{0.282} & \bestres{0.329} & \secondres{0.305} & 0.351 & 0.348 & 0.381 & 0.352 & 0.380 & 0.337 & 0.370 & 0.338 & 0.360 & 0.328 & \secondres{0.346} & 0.341 & 0.349 & 0.515 & 0.481 & 0.410 & 0.387 & 0.396 & 0.379 \\
        \midrule
        \rowcolor{tabhighlight}\multicolumn{2}{c}{\textbf{{$1^{\text{st}}$ Count}}} 
            & \multicolumn{2}{c}{51} & \multicolumn{2}{c}{3} & \multicolumn{2}{c}{1} & \multicolumn{2}{c}{0} & \multicolumn{2}{c}{3} & \multicolumn{2}{c}{0} & \multicolumn{2}{c}{2} & \multicolumn{2}{c}{5} & \multicolumn{2}{c}{1} & \multicolumn{2}{c}{0} & \multicolumn{2}{c}{0} \\
        \bottomrule
        \end{tabular}%
        \begin{tablenotes}
            \item[] $\ast$ Dataset used for pre-training is not evaluated on corresponding models; dashes denote results (--).
            \item[] $\ast$ Traffic from PEMS~\cite{liu2022scinet} is typically used for pre-training large time-series models and is therefore not evaluated here.
        \end{tablenotes}
        \end{threeparttable}
        }
    \end{table}%

    During our experiments, we did not use pre-trained versions of \model, which is left for future work.
    To avoid visual disorder and save space for comparisons with a large set of baselines, we combined the main results of multiple experiments with different versions of \model in \Tabref{tab:results_model_config} and \Tabref{tab:ltsf_full} (see Section~\ref{subsec:forecasting}), presenting them as a single column. 
    However, in \Tabref{tab:results_model_config}, we detail the exact version of \model and the training settings used to achieve the results presented in \Tabref{tab:results_model_config} and \Tabref{tab:ltsf_full}.

    \begin{table}[!htb]
        \caption{Experiment configuration of \model according to the main and additional results reported in Tables \ref{tab:ltsf_avg_full}, \ref{tab:ltsf_full}, and \ref{tab:ltsf_additional}. LR denotes learning rate.}
        \label{tab:results_model_config}
        \centering
        \begin{small}
        \resizebox{0.68\columnwidth}{!}{
        \renewcommand{\multirowsetup}{\centering}
        \setlength{\tabcolsep}{6pt}
        \begin{tabular}{cccccccc}
        \toprule
            \multirow{2}[0]{*}{Dataset} & \multirow{2}[0]{*}{Model version} & \multicolumn{6}{c}{Training Process} \\
        \cmidrule(lr){3-8}
            & & $P$ & $H_o$ & LR & Min LR & Batch Size & Epochs\\
        \midrule
            ETTh1 & \tmodel & 8 & 24 & $3.2 \times 10^{-3}$ & $1.2 \times 10^{-4}$ & 128 & 30 \\
            ETTh2 & \tmodel & 8 & 24 & $3.2 \times 10^{-3}$ & $1.2 \times 10^{-4}$ & 128 & 30 \\
            ETTm1 & \basemodel & 16 & 24 & $3.2 \times 10^{-3}$ & $1.2 \times 10^{-4}$ & 128 & 20 \\
            ETTm2 & \basemodel & 16 & 24 & $3.2 \times 10^{-3}$ & $1.2 \times 10^{-4}$ & 128 & 20 \\
            Weather & \largemodel & 16 & 24 & $3.2 \times 10^{-3}$ & $1.2 \times 10^{-4}$ & 64 & 30 \\
            ECL & \largemodel & 12 & 24 & $2.2 \times 10^{-3}$ & $1.2 \times 10^{-4}$ & 8 & 10 \\
            Traffic & \basemodel & 12 & 24 & $2.2 \times 10^{-3}$ & $1.2 \times 10^{-4}$ & 6 & 15 \\
        \bottomrule
        \end{tabular}
        }
        \end{small}
    \end{table}

\subsection{Patch Lengths}
\label{apdx:patch}

    Forecast performance can be sensitive to patch length $P$. Smaller values increase the sequence length (i.e., the number of patches), reducing computational efficiency and increasing GPU memory demands due to extended sequences. In contrast, larger values may overgeneralize local patterns, degrading accuracy in coarse-grained data.
    Previous works~\cite{zhang2022crossformer,nie2022time,wang2024timexer,liu2024moirai} demonstrate the effects of different patch sizes, suggesting that moderate lengths (e.g., $[8, 24]$) offer a good balance between efficiency and capture of temporal patterns. 
    We evaluate $P \in \{8, 12, 16\}$~\cite{nie2022time}, with \Tabref{tab:forecasting_time} showing our best configurations according to each benchmark. 
    As we can see, for coarse-grained hourly datasets, reduced patch lengths ($P \in \{8, 12\}$) yield superior performances, as they preserve local patterns critical for high-frequency signals.
    On the other hand, minute-level datasets benefit from longer patch lengths ($P = 16$), as larger patches capture broader temporal trends. 
    Increasing or decreasing the length interval resulted in performance degradation, confirming that the optimal patch lengths depend on the dataset frequency~\cite{zhang2022crossformer}.

\subsection{Output Horizons}
\label{apdx:output}

    We evaluate the trade-off between forecasting efficiency and accuracy, ablating \tmodel (\dmodel \(= 64\) and \(P = 8\)) with different output horizons \(H_o \in \{8, 16, 24, 32\}\) on ETTh1 and ETTh2, with MSE and MAE results averaged over the full horizons \(H \in \{96, 192, 336, 720\}\). 
    \Tabref{tab:forecasting_time} presents the MSE, MAE, and total forecasting time for both datasets. Larger \(H_o\) values reduce the number of iterations (that is, \(\lceil H / H_o \rceil\)), enhancing the computational efficiency. 
    For ETTh1, MSE improves from the output horizons \(H_o=8\) to \(H_o=24\) with a $7.0\%$ reduction, before increasing with \(H_o=32\), indicating a possible overgeneralization. 
    Similarly, the ETTh2 columns show a $12.1\%$ reduction in MSE from \(H_o=8\) to \(H_o=24\). 
    The forecast time decreases $69\%$ from the output horizons \(H_o=8\) to \(H_o=24\), but the metrics degrade when \(H_o\) increases to $32$.
    We observed similar behaviors for other benchmark datasets, which confirms \(H_o=24\) as an optimal balance between performance and precision.

    \begin{table}[htb]
        \caption{Ablation study with different output horizons. A lower MSE or MAE indicates a better prediction. The best results are in \textbf{bold}.}
        \label{tab:forecasting_time}
        \centering
        \begin{small}
        \resizebox{0.59\columnwidth}{!}{
        \renewcommand{\multirowsetup}{\centering}
        \setlength{\tabcolsep}{6pt}
        \begin{tabular}{c|c|c|cc|cc|c}
        \toprule
            \multicolumn{1}{c}{Dataset} & \multicolumn{2}{c}{} & \multicolumn{2}{c}{ETTh1} & \multicolumn{2}{c}{ETTh2} &  \\
            \cmidrule(lr){4-5} \cmidrule(lr){6-7}
            Model version & $P$ & $H_o$ & MSE & MAE & MSE & MAE & Time (s) \\
        \midrule
            \multirow{2}{*}[-1.0em]{\begin{tabular}[c]{@{}c@{}}\tmodel\end{tabular}}
            & 8 & 8  & 0.412 & 0.430 & 0.396 & 0.408 & 113 \\
            & 8 & 16 & 0.390 & 0.421 & 0.372 & 0.391 & 57 \\
            & 8 & 24 & \textbf{0.383} & \textbf{0.412} & \textbf{0.348} & \textbf{0.382} & 35 \\
            & 8 & 32 & 0.402 & 0.419 & 0.366 & 0.392 & 26 \\
        \bottomrule
        \end{tabular}
        }
        \end{small}
    \end{table}

\subsection{Scalability Analysis}
\label{apdx:scale_analysis}

    Increasing model size and the number of training tokens generally improves performance, a phenomenon known as scaling laws~\cite{kaplan2020scaling}. 
    We evaluate the impacts of scalability on \model's forecastin performance by varying the representation dimension \(d_{\text{model}} \in \{64, 128, 256, 384\}\) (see \Tabref{tab:model_size}).
    For these experiments, we use the ETTm1, ETTm2, Weather, and ECL datasets. 
    ETTh1 and ETTh2 were excluded due to the high risks of overfitting on such small datasets~\cite{nie2022time}, as well as Traffic, due to hardware constraints. 

    \begin{figure}[!htb]
        \centering
        \includegraphics[width=1\columnwidth]{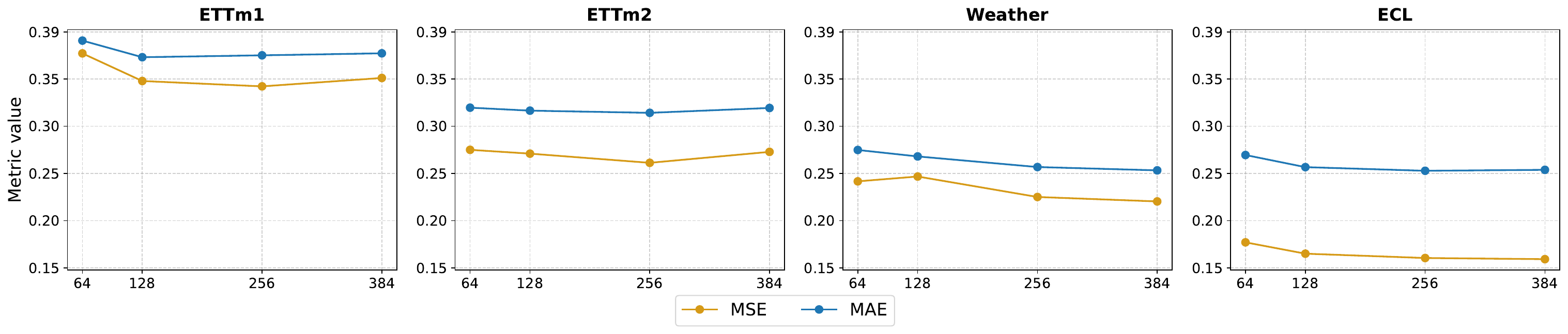}
        \caption{Scalability analysis on ETTm1, ETTm2, Weather, and ECL, with varying \dmodel sizes on the x-axis. Lower MSE or MAE indicates better performance.}
        \label{fig:scalability}
    \end{figure}
    
    Figure~\ref{fig:scalability} shows that increasing the size of \model improves both MSE and MAE, confirming scalability benefits in the time-series domain~\cite{shi2024time}. 
    On ETTm1 and ETTm2, MSE decreases substantially up to \dmodel $ = 256$, then increases slightly at \dmodel $ = 384$, suggesting potential overfitting due to the limited data size. 
    In contrast, \model achieves consistent reductions in metrics up to \dmodel $ = 384$ on Weather and ECL, indicating robust scaling benefits on larger datasets.
    These results not only align with scaling laws but also validate \model's potential for further scaling in resource-rich contexts.

\section{Forecast Showcases}
\label{apdx:showcases}

\begin{figure}[!htb]
        \centering
        \vspace{6pt}
        \includegraphics[width=1\columnwidth]{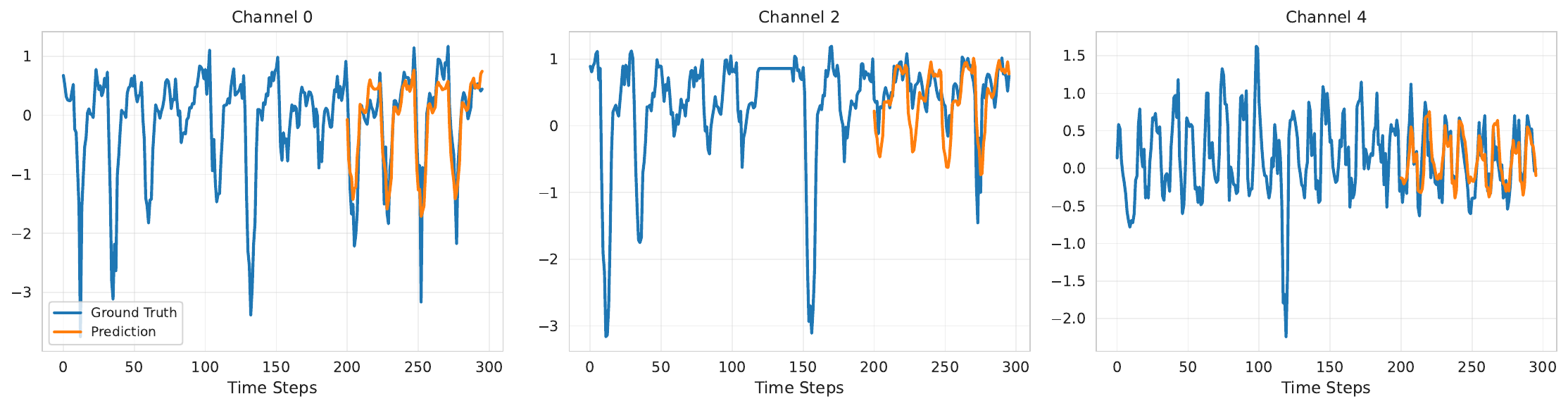}
        \caption{Forecast showcases of \model across different time channels from \textbf{ETTh1}, with a horizon of 96. Blue curves are the ground truths, and orange curves are the model predictions. Curves before the model predictions are the input data.}
        \label{fig:show_etth1}
    \end{figure}

\begin{figure}[!htb]
        \centering
        \vspace{6pt}
        \includegraphics[width=1\columnwidth]{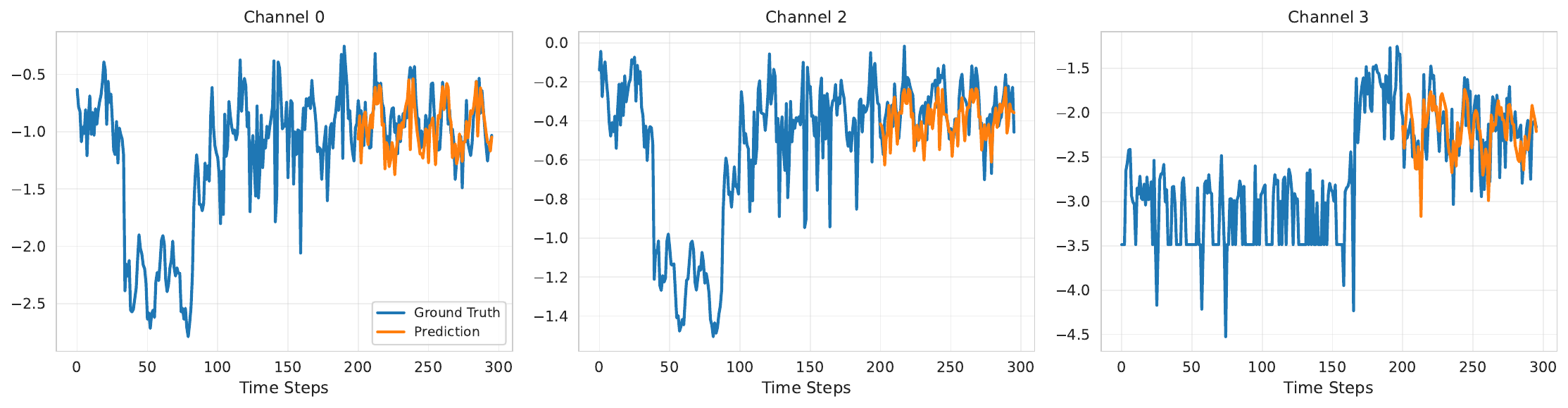}
        \caption{Forecast showcases of \model across different time channels from \textbf{ETTh2}, with a horizon of 96. Blue curves are the ground truths, and orange curves are the model predictions. Curves before the model predictions are the input data.}
        \label{fig:show_etth2}
    \end{figure}

\begin{figure}[!htb]
        \centering
        \vspace{6pt}
        \includegraphics[width=1\columnwidth]{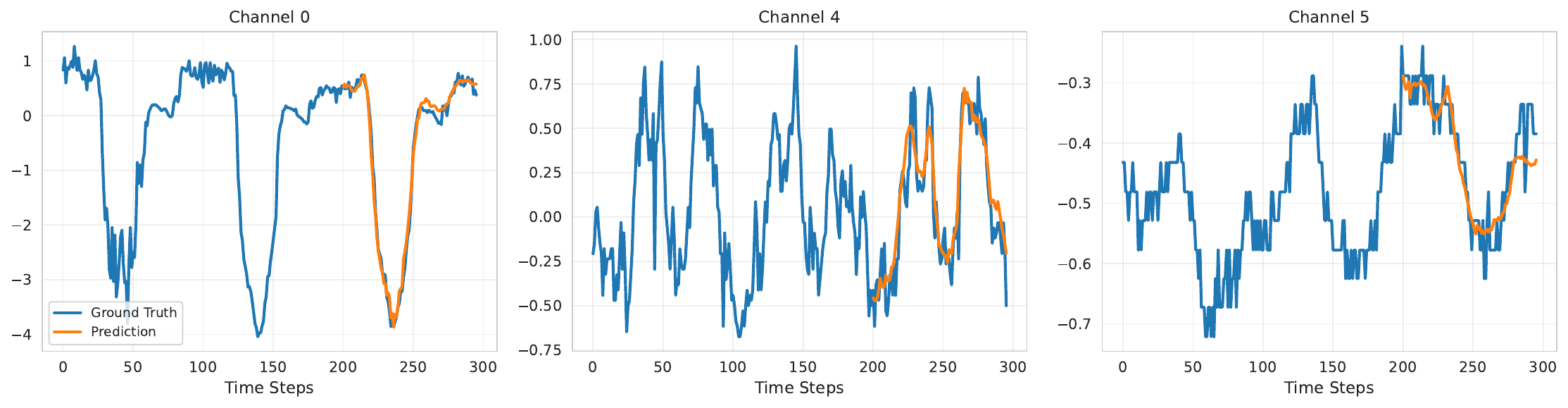}
        \caption{Forecast showcases of \model across different time channels from \textbf{ETTm1}, with a horizon of 96. Blue curves are the ground truths, and orange curves are the model predictions. Curves before the model predictions are the input data.}
        \label{fig:show_ettm1}
    \end{figure}

\begin{figure}[!htb]
        \centering
        \vspace{6pt}
        \includegraphics[width=1\columnwidth]{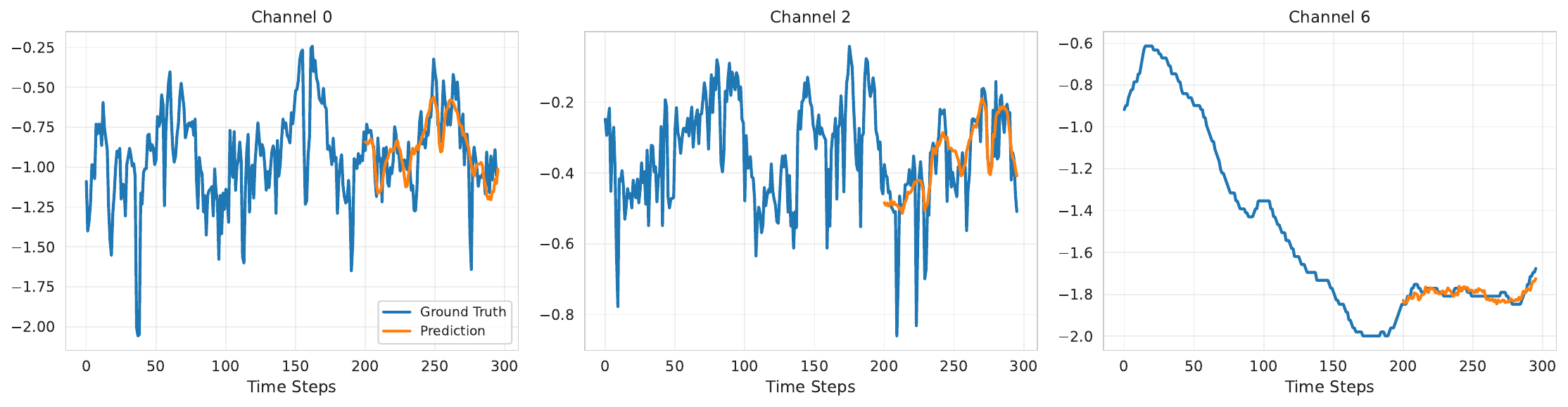}
        \caption{Forecast showcases of \model across different time channels from \textbf{ETTm2}, with a horizon of 96. Blue curves are the ground truths, and orange curves are the model predictions. Curves before the model predictions are the input data.}
        \label{fig:show_ettm2}
    \end{figure}

\begin{figure}[!htb]
        \centering
        \vspace{6pt}
        \includegraphics[width=1\columnwidth]{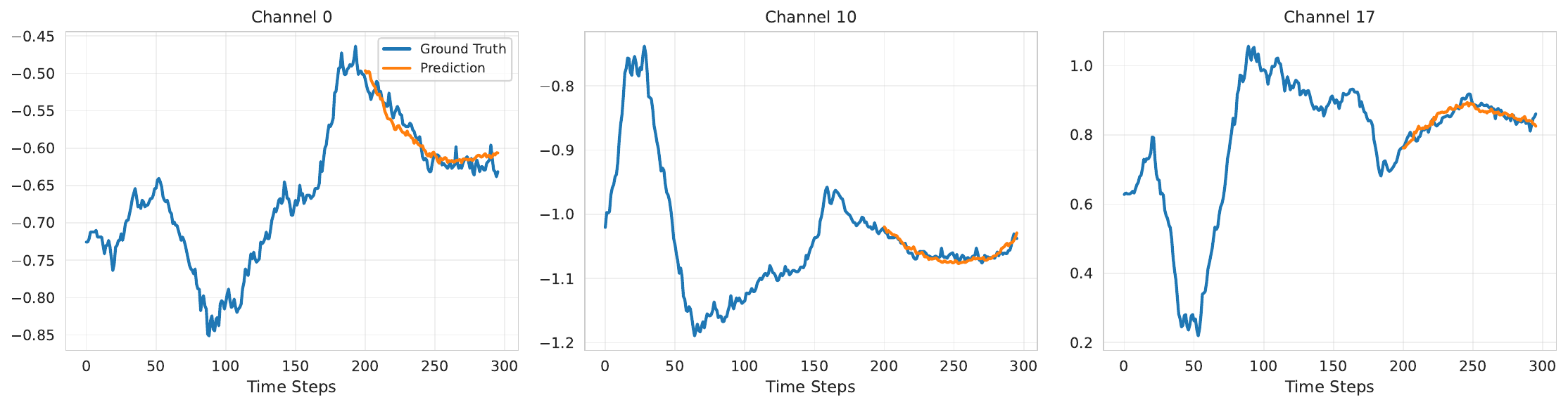}
        \caption{Forecast showcases of \model across different time channels from \textbf{Weather}, with a horizon of 96. Blue curves are the ground truths, and orange curves are the model predictions. Curves before the model predictions are the input data.}
        \label{fig:show_weather}
    \end{figure}

\begin{figure}[!htb]
        \centering
        \vspace{6pt}
        \includegraphics[width=1\columnwidth]{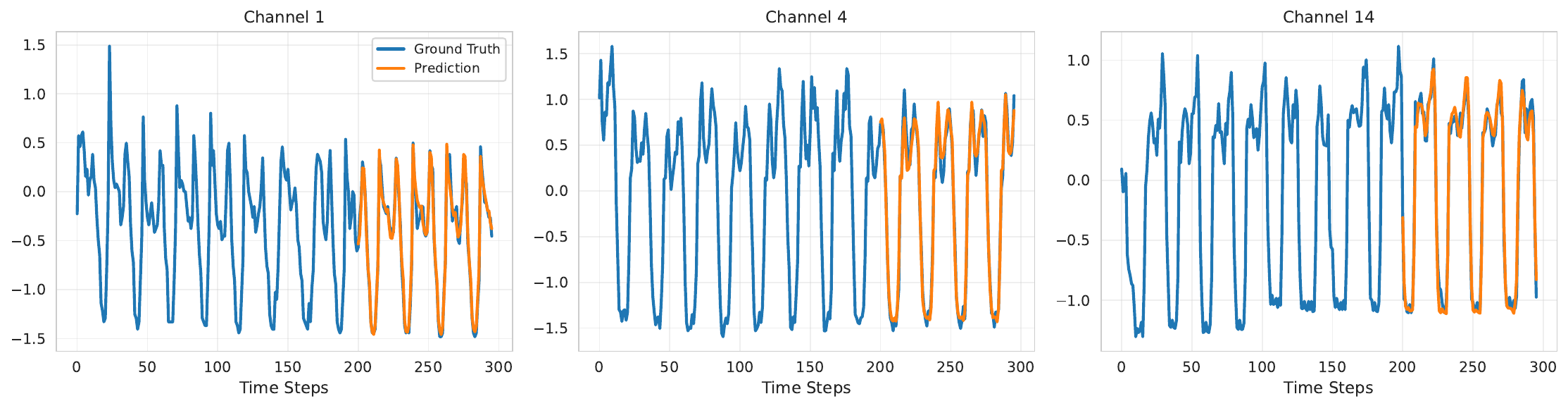}
        \caption{Forecast showcases of \model across different time channels from \textbf{ECL}, with a horizon of 96. Blue curves are the ground truths, and orange curves are the model predictions. Curves before the model predictions are the input data.}
        \label{fig:show_ecl}
    \end{figure}

\begin{figure}[!htb]
        \centering
        \vspace{6pt}
        \includegraphics[width=1\columnwidth]{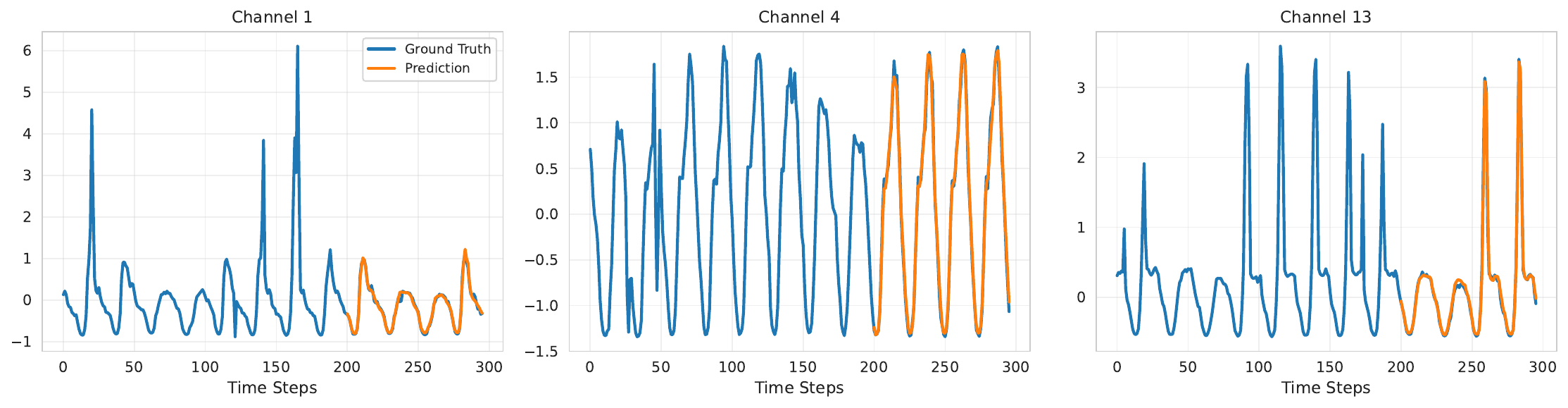}
        \caption{Forecast showcases of \model across different time channels from \textbf{Traffic}, with a horizon of 96. Blue curves are the ground truths, and orange curves are the model predictions. Curves before the model predictions are the input data.}
        \label{fig:show_traffic}
    \end{figure}

    To provide a qualitative assessment of \model's performance, we visualize its forecasting results across different time dimensions from the test sets of the benchmark datasets, namely ETTh1, ETTh2, ETTm1, ETTm2, Weather, ECL, and Traffic (Figures~\ref{fig:show_etth1}~--~\ref{fig:show_traffic}). In each figure, the forecast horizon is set to 96 time steps. To enhance clarity and ensure intuitive visualizations, we display the full predicted horizon alongside a slice of the historical input data (look-back window) and the corresponding ground-truth future values.

    These visualizations illustrate \model's ability to generate accurate and coherent forecasts in highly heterogeneous multivariate time series, underscoring the effectiveness of the proposed architecture. As demonstrated in quantitative evaluations (see Section~\ref{apdx:results}), \model's performance gains are particularly pronounced in long- and ultra-long-term prediction settings, where it robustly captures complex temporal dynamics. Overall, these qualitative results highlight the practical utility of \model for state-of-the-art, long-term multivariate time series forecasting.

\section{Discussion, Limitations, and Future Work}
\label{apdx:limitations_future}

    Although both language and time series are sequential data with long-range dependencies, they differ fundamentally: language relies on deterministic structures and semantic patterns, whereas time series data often arise from stochastic processes with complex temporal dynamics. Time is not language. Consequently, effective temporal modeling requires specialized architectures beyond scaling language models. \model introduces tailored innovations that combine multiple approaches, respecting temporal structures such as time continuity, seasonality, and non-stationarity.
    
    Although \model demonstrates significant capabilities, specific directions warrant further exploration. 
    In contexts with large model configurations (e.g., \dmodel $= 384$) and high-dimensional datasets with numerous features, inference latency increases, particularly for ultra-long horizons (e.g., \(H = 720\) time points). 
    Increasing the output horizon significantly improves forecasting; however, long output horizons can degrade accuracy, especially on coarse-grained datasets, due to overgeneralization of temporal patterns. 
    During our experiments, we identified \(H_o = 24\) as an optimal balance between inference speed and precision on diverse benchmarks, outperforming standard ``next-token'' autoregressive predictions, because it allows us to generate $24\times$ more future time points per iteration.
    
    Caching mechanisms could be used to speed up forecasting. Widely used in language models to accelerate inference, KV Caching~\cite{pope2023efficiently} has not been thoroughly investigated in time-series forecasting, particularly for patch-level attention encoders with multi-resolution outputs. 
    Thus, adapting caching approaches to encoder-only models is a promising direction to reduce latency for long-horizon predictions.
    Another inspiring concept for future exploration is the multi-resolution projection head~\cite{shi2024time}, in which multiple projection heads, each corresponding to a distinct forecasting horizon, are optimized during training to provide more flexibility for handling time-series data with different frequencies.

    Enhancing covariate integration through adaptive mechanisms to handle missing or temporally misaligned exogenous covariates could improve robustness in real-world scenarios~\cite{wang2024timexer}. 
    Furthermore, extending the Mixture-of-Heterogeneous-Experts (MoHE) to incorporate additional architectures (e.g., graph-based or attention-based experts) could capture more complex temporal dependencies, broadening \model's applicability to heterogeneous time-series tasks.
    Finally, scale and pre-train \model on large-scale time-series datasets, such as Time-MoE~\cite{shi2024time} and TimesFM~\cite{das2023decoder}, could enable zero-shot forecasting across diverse domains, leveraging the MoHE architecture's sparsity for efficient large-scale training.

\end{document}